\documentclass[10pt,twocolumn,letterpaper]{article}

\usepackage{cvpr}              % To produce the CAMERA-READY version
\usepackage[table]{xcolor}
\usepackage{multicol,multirow,tabularx}
\usepackage{booktabs}
\usepackage{colortbl}
\usepackage{graphicx}
\usepackage{float} % Add this to your preamble
\usepackage{makecell}
\usepackage{cuted}   % provides 'strip'
\usepackage{url}
\usepackage{soul} 
\usepackage{fontawesome5}
\newcolumntype{Y}{>{\raggedright\arraybackslash}X}

\newcommand{\myparagraph}[1]{\vspace{0.2em}\noindent\textbf{#1}}

\definecolor{cvprblue}{rgb}{0.21,0.49,0.74}
\usepackage[pagebackref,breaklinks,colorlinks,allcolors=cvprblue]{hyperref}

\usepackage{helvet}

\title{%
\vspace*{-15mm}
\begin{minipage}{\textwidth}
    \includegraphics[height=0.5cm]{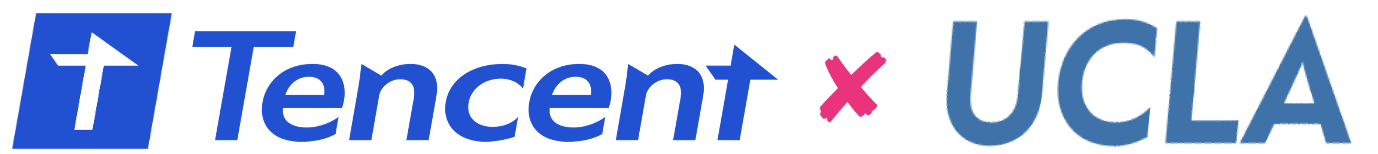}\\[-12pt]
    \noindent\rule{\textwidth}{0.8pt}
\end{minipage}\\[30pt] % space between header and actual title
\sffamily\bfseries\fontsize{14}{22}\selectfont
MotionEdit: Benchmarking and Learning Motion-Centric Image Editing
}

\author{Yixin Wan$^{1,2}$\textsuperscript{\thanks{Work done during internship at Tencent AI Lab in Seattle, contact email: \texttt{elaine1wan@cs.ucla.edu}}}, Lei Ke$^1$, Wenhao Yu$^1$, Kai-Wei Chang$^2$, Dong Yu$^1$ 
\\
$^1$Tencent AI, Seattle \ 
$^2$University of California, Los Angeles
\\
\\
\faGithub \ Project Page: \href{https://motion-edit.github.io/}{https://motion-edit.github.io}
}

\begin{document}
\maketitle

\begin{strip}
\centering
\vspace{-12mm}
\includegraphics[width=\textwidth]{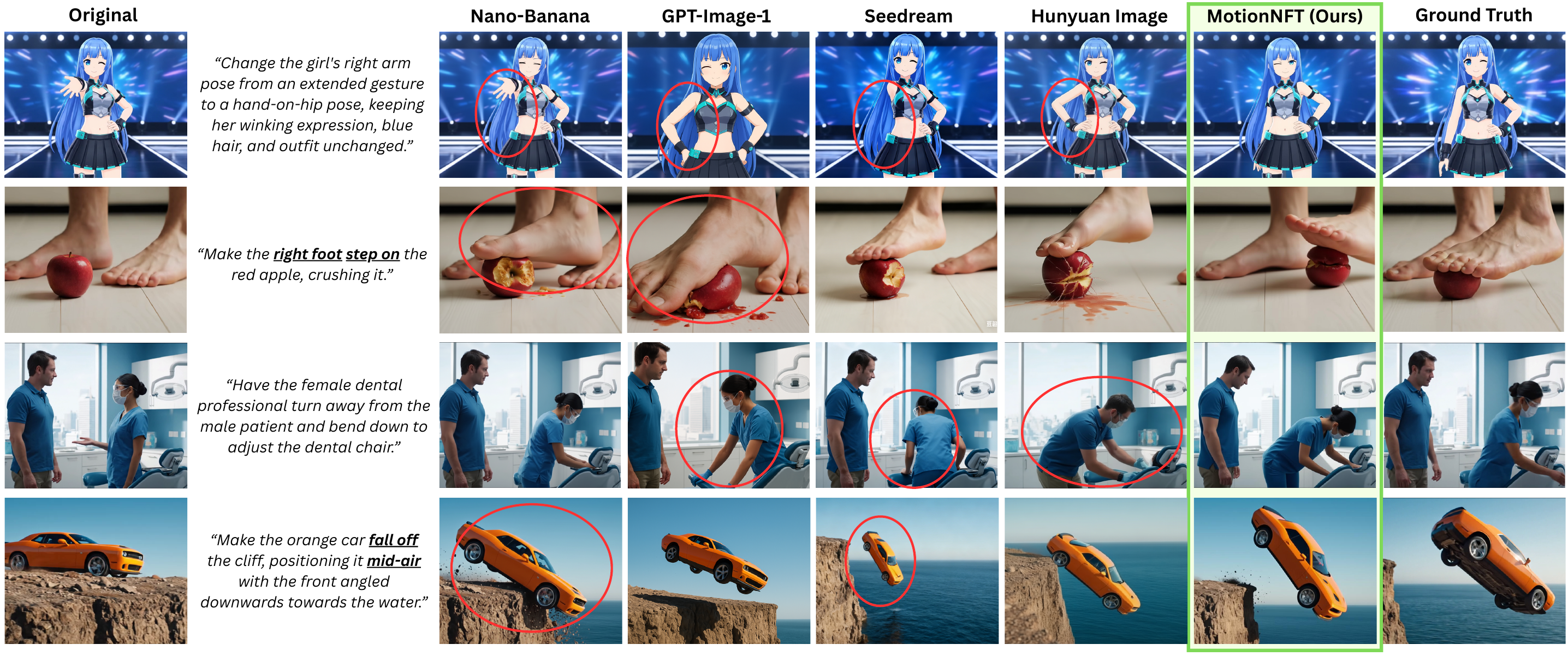}
\vspace{-6mm}
\captionof{figure}{\label{fig:fail-closed-example}Case studies of MotionNFT against closed-source commercial baselines: Nano-Banana~\citep{nanobanana}, GPT-Image-1~\citep{openai2025imagegen}, Seedream~\citep{seedream2025seedream}, and Hunyuan Image~\citep{cao2025hunyuanimage}. Red circles highlight failure regions (e.g., failing to displace the car in the bottom row, generating an artifact `foot'' in the second row). While commercial models maintain high visual quality, they struggle to ground complex motion changes or maintain visual consistency. MotionNFT follows these dynamic instructions, ensuring geometric alignment with the ground truth.}
\end{strip}
% either lack meaningful action or motion edits. They e

% Comparison of existing image-editing benchmarks with \textbf{\textsc{MotionEdit}}. Prior datasets either focus solely on appearance adjustments, or only contain low-quality, incoherent action edits (e.g. InstructP2P~\citep{brooks2022instructpix2pix}'s target edited image does not follow the instruction with multiple artifact distortions). In contrast, our \textbf{\textsc{MotionEdit}} provides high-quality, paired input–edit examples spanning rich motion and interactions.

\begin{abstract}
We introduce \textbf{MotionEdit}, a novel dataset for motion-centric image editing—the task of modifying subject actions and interactions while preserving identity, structure, and physical plausibility.
Unlike existing image editing datasets that focus on static appearance changes or contain only sparse, low-quality motion edits, MotionEdit provides high-fidelity image pairs depicting realistic motion transformations extracted and verified from continuous videos. This new task is not only scientifically challenging but also practically significant, powering downstream applications such as frame-controlled video synthesis and animation.

To evaluate model performance on the novel task, we introduce \textbf{MotionEdit-Bench}, a benchmark that challenges models on motion-centric edits and measures model performance with generative, discriminative, and preference-based metrics.
Benchmark results reveal that motion editing remains highly challenging for existing state-of-the-art diffusion-based editing models.
To address this gap, we propose \textbf{MotionNFT} (Motion-guided Negative-aware FineTuning), a post-training framework that computes motion alignment rewards based on how well the motion flow between input and model-edited images matches the ground-truth motion, guiding models toward accurate motion transformations.
Extensive experiments on FLUX.1 Kontext and Qwen-Image-Edit show that MotionNFT consistently improves editing quality and motion fidelity of both base models on the motion editing task without sacrificing general editing ability, demonstrating its effectiveness.

\begin{figure*}[thb]
\centering
\vspace{-6mm}
\includegraphics[width=\textwidth]{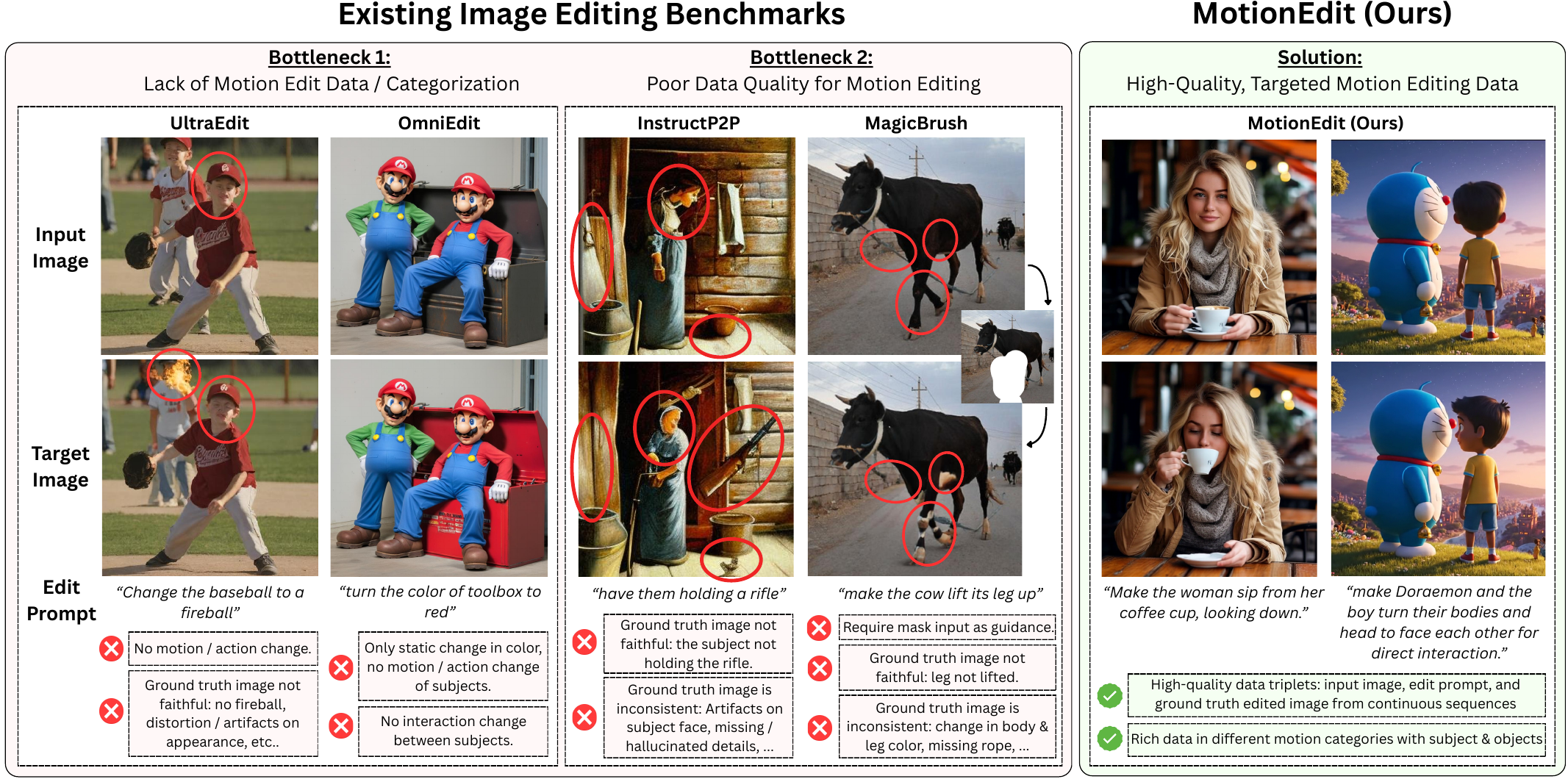}
\vspace{-5mm}
\captionof{figure}{Comparison of existing image editing benchmarks with \textbf{\textsc{MotionEdit}}.
Prior datasets lack motion-edit supervision, either focusing on appearance edits or offering low-quality action changes with artifacts.
\textbf{\textsc{MotionEdit}} fills this gap by providing instruction-following motion edits with paired input–target image data, enabling evaluation and training of motion-aware image editing models.}
\label{fig:benchmark-comparison}
\vspace{-0.3em}
\end{figure*}

\end{abstract}

\vspace{-6mm}
\section{Introduction}
Instruction-guided image editing models have made remarkable progress recently~\citep{labs2025flux1kontextflowmatching,wu2025qwenimagetechnicalreport,deng2025bagel,openai2025imagegen,lin2025uniworld}, capable of transforming images based on natural language commands.
While recent image editing models excel at performing appearance-only static edits that simply adjust color, texture, or object presence, they oftentimes fall short in accurately, faithfully, and naturally editing the motion, posture, or interaction between subjects in images.
In this work, we aim at addressing this limitation in existing models through systematically formulating and studying motion editing as an independent and important image editing task. % improving existing models' capabilities ...?

\begin{figure*}[thb]
\centering
% \vspace{-6mm}
\includegraphics[width=0.96\textwidth]{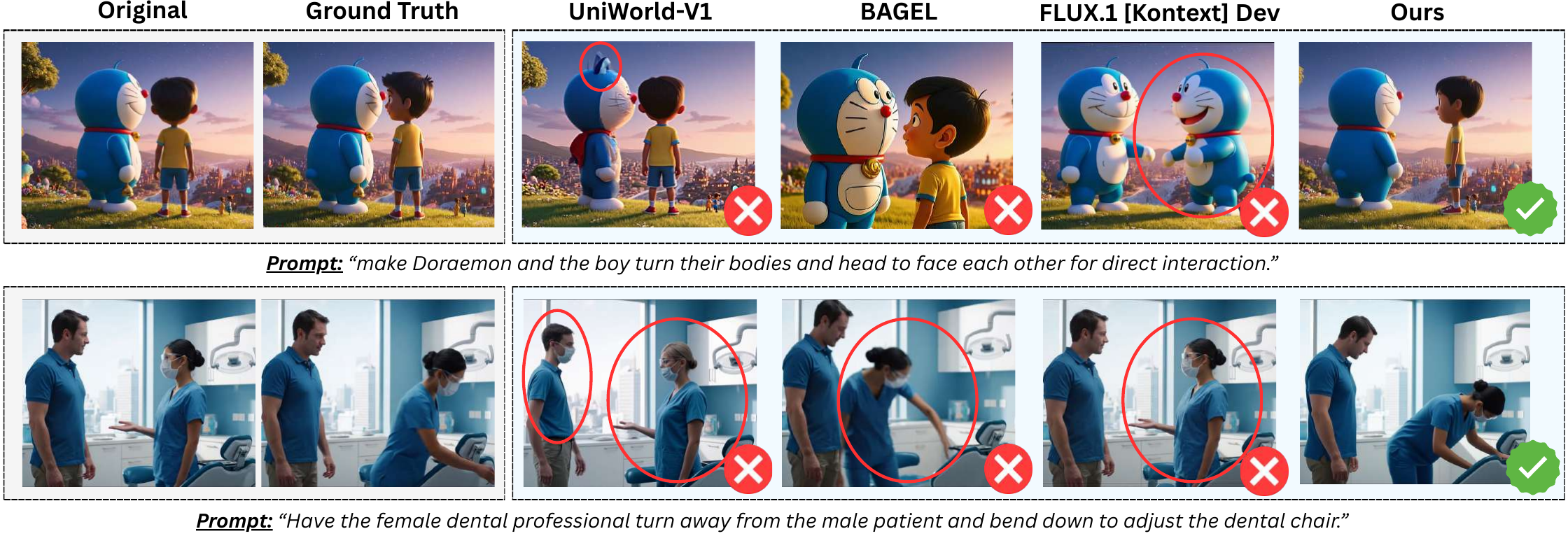}
\vspace{-0.2em}
\captionof{figure}{\label{fig:model-comparison}Qualitative comparison of state-of-the-art image editing models on \textsc{MotionEdit}. 
Existing models fail to execute the required motion edits (e.g. \textsc{UniWorld-V1} fail to edit subject postures and \textsc{FLUX.1 Kontext} produces severe identity distortions), while our MotionNFT-trained model accurately performs the intended motion edit that closely matches the ground-truth.}
\vspace{-0.3em}
\end{figure*}
We formally define the new task of \textbf{motion image editing}---editing that modifies the action, pose, or interaction of subjects and objects in an image according to a textual instruction, while preserving visual consistency in characters and scene.
Motion editing aims at changing \textit{how} subjects move, act, or interact, which is essential for applications such as frame-controlled video generation and character animation.
% , and interactive content creation.
However, existing image editing datasets and benchmarks suffer from two major bottlenecks in approaching the motion image editing task: 
First, they primarily focus on static editing tasks like appearance modification or replacement (e.g. OmniEdit~\citep{wei2024omniedit} and UltraEdit~\citep{zhao2024ultraeditinstructionbasedfinegrainedimage} examples in Figure \ref{fig:benchmark-comparison}), neglecting the important aspect of motion editing in their data at all.
Second, datasets that do include motion edits offer only a small amount of low-quality data, often with unfaithful or incoherent edit ground-truth that fail to execute the intended motion (e.g. InstructP2P~\citep{brooks2022instructpix2pix} and MagicBrush~\citep{Zhang2023MagicBrush} examples in Figure~\ref{fig:benchmark-comparison}).

To bridge this research gap, we curate \textbf{\textsc{MotionEdit}}, a high-quality dataset and benchmark specifically targeting motion editing, consisting of paired input–target image examples extracted and validated from continuous high-resolution video frames to ensure accurate, natural, and coherent motion changes.
As shown in Figure \ref{fig:benchmark-comparison}, \textsc{MotionEdit} captures realistic action and interaction changes that preserve identity, background, and style, in contrast to prior datasets where edit data is either static, unfaithful, or visually inconsistent.
Moreover, our data is sourced from a large set diverse video sequences, ensuring the assessment of diverse sub-categories of motion image editing, such as posture, orientation, and interaction changes in Figure \ref{fig:benchmark-examples}.
Beyond constructing high quality editing data, we also devise evaluation metrics to evaluate motion edit performances of models.
% This flow-based metric provides a geometry-aware, discriminative signal for motion correctness, and serves as a strong complement to 
For discriminative evaluation, we by comparing the optical flow~\citep{sun2018pwc,teed2020raft,jiang2021learning,xu2022gmflow,xu2023unifying}---which captures the magnitude and direction of motion change---between the input and model-edited images against the input–ground truth flow.
For generative evaluation, we adopt Multimodal Large Language Model (MLLM)-based metrics to assess the fidelity, preservation, coherence, and overall quality of edited images.
Additionally, we report pairwise win rates through head-to-head comparisons between overall edit quality of different models to reflect preference performance.
% We evaluate 9 state-of-the-art image editing models using \textsc{\textbf{MotionEdit-Bench}}.
Both quantitative and qualitative results across state-of-the-art image editing models on \textsc{MotionEdit-Bench} show that \textbf{motion image editing remains a challenging task for the majority of open-source image editing models.}
% , providing quantitative measurements and qualitative analysis on their motion editing performances.

% We further explore the potential of optical flow-based motion estimation t
To improve existing image editing models on the motion editing task, we further propose Motion-guided Negative-aware FineTuning (\textbf{\textsc{MotionNFT}}), a post-training framework for motion editing that extends DiffusionNFT~\citep{zheng2025diffusionnft} to incorporate motion-aware reward signals. 
MotionNFT leverages the motion alignment measurement between input-edit and input-ground truth optical flows to construct a reward scoring framework, providing targeted guidance on motion direction and magnitude in training.
As illustrated in Figure~\ref{fig:model-comparison}, MotionNFT enables models to perform accurate, geometrically consistent motion edits.
Quantitative results in Table \ref{tab:edit_comparison} further shows substantial improvement across all metrics over prior approaches.
For instance, \textsc{MotionNFT} achieves over 10\% improvement in overall quality and over 12\% on pairwise win rates when applied on FLUX.1 Kontext~\citep{labs2025flux1kontextflowmatching}).

The key contributions of our paper are three-fold:
\begin{itemize}
    \item We systematically define and study the novel task of \textbf{motion image editing}.
    \item We construct \textbf{\textsc{MotionEdit}}, a high quality dataset and benchmark for motion image editing, containing diverse and accurate edit data sourced from video frames.
    \item We propose \textbf{\textsc{MotionNFT}}, a post-training framework that integrates optical flow–based rewards into DiffusionNFT to guide motion edit improvements.\footnote{Dataset, code, and evaluation toolkit will be released upon acceptance.}
\end{itemize}

\section{Related Works}
\label{sec:related-work}

\noindent\textbf{Image Editing.} Recent advances in text-to-image (T2I) diffusion models have greatly improved text-guided image editing~\citep{mengsdedit,brooks2022instructpix2pix,Zhang2023MagicBrush,labs2025flux1kontextflowmatching,wu2025qwenimagetechnicalreport,openai2025imagegen}. While current models handle static appearance edits well (e.g., color changes or object replacement), they struggle with motion-related edits that require modifying actions or interactions (e.g., ``make the man drink from the cup'').
This gap largely stems from limitations in existing editing datasets. First, most benchmarks focus on static transformations—local texture changes, object replacement, or style transfer~\citep{zhao2024ultraeditinstructionbasedfinegrainedimage,wei2024omniedit,brooks2022instructpix2pix}—with little coverage of motion edits. Second, datasets containing motion edits are small and of low quality: motion categories are unclear, and the provided target edits are often unfaithful or physically implausible~\citep{brooks2022instructpix2pix,Zhang2023MagicBrush,lin2024schedule}. As shown in Fig.~\ref{fig:benchmark-comparison}, these models frequently fail to achieve intended action changes and introduce visual artifacts, undermining both training supervision and evaluation reliability.
These limitations underscore a key challenge in motion image editing: building datasets with precise motion-edit instructions and high-quality, faithful edited targets that preserve appearance and scene context while accurately reflecting the intended action changes.

\noindent\textbf{Motion Estimation in Images.}
Motion estimation is a long-standing problem in computer vision.
% Early methods such as block matching~\citep{cuevas2013block} estimate motion by searching for similar patches across frames. 
Modern approaches rely on optical flow, which predicts per-pixel displacement between two images~\citep{sun2018pwc, teed2020raft,jiang2021learning,xu2022gmflow}. 
Recent work such as UniMatch~\citep{xu2023unifying} further advances large-displacement estimation by formulating optical flow as a global matching problem unified with stereo tasks.
Inspired by the effectiveness of optical flow in capturing fine-grained motion changes, we propose a motion-centric reward framework based on optical flow, which quantitatively measures how accurately a model performs the intended motion edit in synthesized images.

\noindent\textbf{Reinforcement Learning for Image Generation.}
Policy-gradient methods such as PPO~\citep{schulman2017proximalpolicyoptimizationalgorithms,ren2024diffusion} and GRPO~\citep{shao2024deepseekmath,liu2025flow} have been explored for improving image generation. 
More recently, DiffusionNFT~\citep{zheng2025diffusionnft} introduces negative-aware finetuning, which contrasts positive and negative generations during the forward diffusion process to obtain an implicit policy improvement direction, steering the model toward high-reward outcomes while repelling low-reward ones.
UniWorld-V2~\citep{lin2025uniworld} extends DiffusionNFT by integrating an MLLM-based online scoring pipeline for rating editing aspects like prompt compliance and style fidelity.  
However, current RL-based post-training frameworks remain motion-agnostic: they emphasize semantic correctness and visual details, yet offer no supervision on how subjects and objects should \emph{move} for motion-centric edits.

\section{Dataset Construction}
% Our contribution: the benchmark and the training data.

\begin{figure*}[t!]
% \vspace{-0.6em}
 \centering
\includegraphics[width=\textwidth]{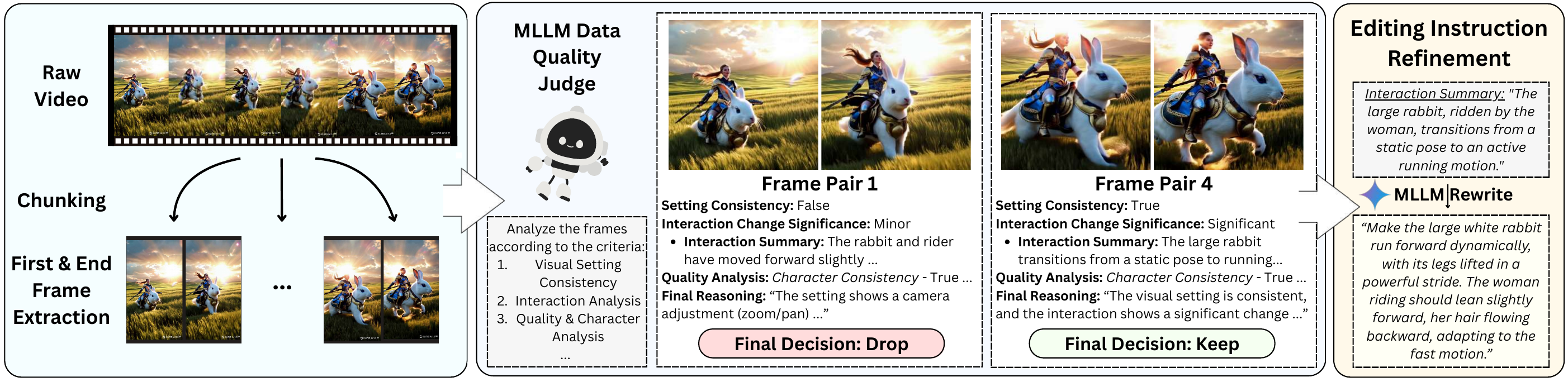}
  \vspace{-1em}
\caption{\label{fig:benchmark-construction}MotionEdit's data construction pipeline. We segment raw videos, extract frame pairs, and automatically filter them using an MLLM data quality judge. For all kept pairs, we use a MLLM rewrite module to generate clean, motion-focused editing instructions. Our pipeline enables scalable construction of high-quality motion editing data and can be extended to much larger video corpora.}
 \vspace{1.2em}
\includegraphics[width=\textwidth]{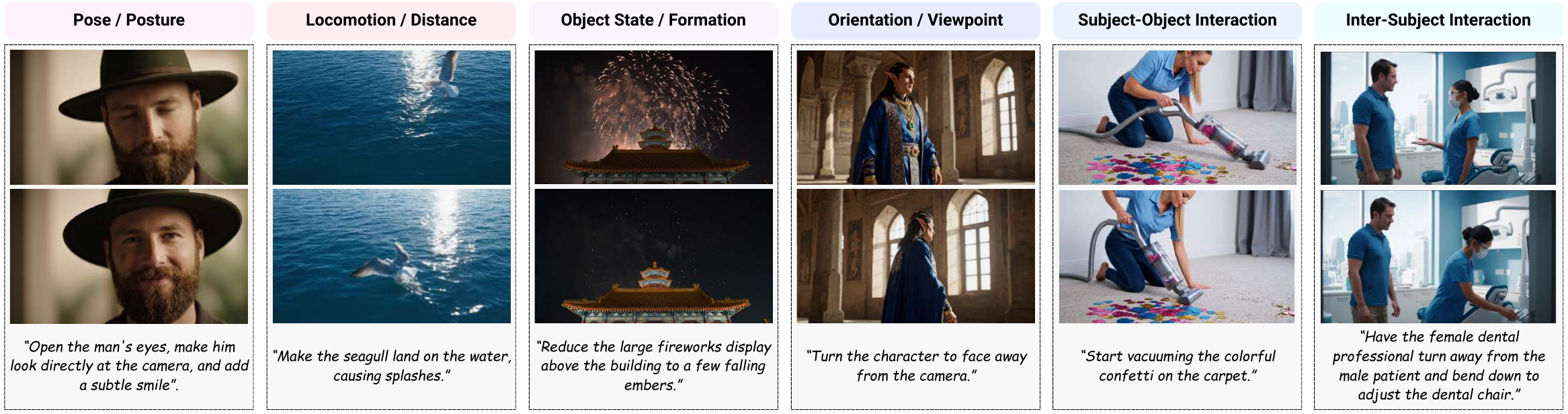}
 \vspace{-1em}
\caption{\label{fig:benchmark-examples} Example categories of data in \textsc{MotionEdit}. Drawing from diverse video sources, our dataset captures a broad spectrum of motion transformations, including pose shifts, locomotion, viewpoint changes, and both subject–object and inter-subject interactions.}
% \vspace{-0.8em}
\end{figure*}

\subsection{Problem Definition and Categorization}
The task of motion image editing has not been comprehensively explored in prior works.
Therefore, we first provide a systematic definition of this novel task.

\myparagraph{Motion Image Editing.}
Given an input image and a natural-language instruction specifying a target motion change (e.g. \textit{``make the woman drink from the cup''}), the goal is to synthesize an edited image where:  
(1) the edited motion faithfully reflects the intended action; 
(2) the resulting pose or interaction is physically plausible and respects articulated constraints (e.g., \textit{``slightly open his eyes''});
(3) non-edited factors like appearance, background, and viewpoint remains consistent.  
Unlike traditional appearance-focused editing, motion editing requires models to interpret the instructed motion and translate it into coherent spatial changes in the image, requiring fine-grained spatial and kinematic understanding.

\subsection{Dataset Construction Pipeline}
As discussed in Section~\ref{sec:related-work}, existing image editing datasets and benchmarks lack reliable ground-truth targets that correctly execute the instructed motion while preserving subject identity and scene context. Prior datasets either introduce artifacts and hallucinations, alter appearance, or unintentionally shift viewpoint or scale. 
Sourcing high-quality motion edit ground truth remains a challenging problem.
% This raises a fundamental question: \emph{Where do we obtain truly correct ``moved pixels''?}
% We highlight that ground truth motion edits naturally occur in video sequences with movements. 
Instead of synthesizing edited targets as in prior work~\citep{brooks2022instructpix2pix,Zhang2023MagicBrush}, we propose a \textbf{video-driven data construction pipeline} that mines paired frames from dynamic video sequences to produce high-quality (input image, edit instruction, target image) triplets. These data reflect naturally occurring and coherent motion transitions grounded in video kinematics.
Full details on dataset construction are in the ``Additional Dataset Construction Details'' Appendix section.

\myparagraph{Video Collection}
To obtain frame pairs capturing clean motion transitions, we first explored conventional human action datasets such as HAA500~\citep{haa500} and K400~\citep{kay2017kinetics}. 
Although diverse, these datasets often suffer from problems like low resolution, motion blur, rapid viewpoint shifts, etc., making them unsuitable for extracting faithful pre-/post-edit pairs that preserve identity and background consistency.

In contrast, recent Text-to-Video (T2V) models (e.g. Veo-3~\citep{veo3_2025}, Kling-AI~\citep{klingai_2025}) produce visually sharp, temporally smooth videos with stable subjects and backgrounds. We therefore draw from two publicly released T2V video collections---ShareVeo3~\citep{wang2024vidprom} and the KlingAI Video Dataset~\citep{klingai_dataset_2025}---as our initial pool of candidate videos. 
We then apply further processing to extract high-quality frame pairs for our \textsc{MotionEdit} dataset.

\myparagraph{Frame Extraction and Automatic Validation}
Given the video pool, our goal is to identify frame pairs that exhibit meaningful motion changes while preserving all non-motion factors. We segment each video into 3-second windows and sample the first and last frame of each segment, providing a broad and efficient set of candidate motion transitions.
However, many sampled pairs are unusable due to camera motion, subject disappearance, environmental changes, or visual degradation. 
Motivated by the recent success of LLM/MLLM-based data filtering~\citep{chen2024your,henriksson2025finerweb,wang2025open,cayir2025refine}, we leverage Google's Gemini~\citep{team2023gemini} model to automatically filter these cases at scale. We prompt Gemini to evaluate each frame pair along three critical dimensions:

\begin{itemize}
    \item \textbf{Setting Consistency.} Verify that background, viewpoint, and lighting remain stable despite subject motion.
    \item \textbf{Motion and Interaction Change.} Identify interaction states in each frame and summarize the primary motion transition (e.g., ``not holding cup $\rightarrow$ drinking''). The model also judges whether the change is significant enough to constitute a meaningful motion edit.
    \item \textbf{Subject Integrity and Quality.} Ensure the main subjects are present, identifiable, and artifact-free, avoiding cases with occlusion, shrinkage, hallucinations, and distortions.
\end{itemize}

Based on these criteria, the MLLM outputs a binary keep/discard decision. A pair is accepted only if (1) the scene remains stable, (2) the motion change is non-trivial, (3) subjects are consistent and coherent, and (4) both frames maintain high visual quality. This filtering is essential for obtaining high-quality motion edit triplets for our dataset.

\subsection{Editing Prompt Construction}
While the validated frame pairs provide reliable visual reference, their corresponding edit instructions must be clear, natural, and semantically faithful to the observed change. We convert the MLLM-generated motion-change summaries into user-style editing prompts by following the prompt refinement procedure of \citet{wu2025qwenimagetechnicalreport}. This step removes unnecessary analysis details and standardizes prompts into imperative form (e.g. ``Make the woman turn her head toward the dog.''), ensuring alignment between the described edit and the actual motion transition in data.

\subsection{Dataset Statistics}
Our final \textsc{MotionEdit} dataset consists of 10,157 motion-editable frame pairs, sourced from both Veo-3 and KlingAI video collections. 
Specifically, we obtain 6,006 samples from Veo-3 and 4,151 samples from KlingAI. 
We perform a random 90/10 train-test split, resulting in 9,142 training data and 1,015 evaluation data that constitutest \textsc{MotionEdit-Bench}. 
Each sample includes a source or input image, a target image exhibiting a real motion transition from the original video, and a precise motion edit instruction.
As shown in Figure \ref{fig:benchmark-examples}, data in \textsc{MotionEdit} can be generally categorized into six motion edit types:
\begin{itemize}
    \item \textbf{Pose / Posture}: Changes in body configuration position (e.g. raising hand) while keeping identity and scene fixed.
    \item \textbf{Locomotion / Distance}: Changes in subject’s spatial position or distance relative to the camera or environment.
    \item \textbf{Object State / Formation}: Changes in the physical form or condition of an object (e.g., deformation, expansion).
    \item \textbf{Orientation / Viewpoint}: Changes in subject’s facing direction or angle without position change.
    \item \textbf{Subject–Object Interaction}: Changes in how a person or agent physically interacts with an object (e.g., holding).
    \item \textbf{Inter-Subject Interaction}: Changes in the coordinated motion between two or more subjects (e.g., facing).
\end{itemize}

\subsection{Data Motion Magnitude Comparison}
To quantify and compare the amount of motion present in before-after editing pairs between \textsc{MotionEdit} and other editing datasets, we randomly select 100 data from each dataset and calculate the overall pixel-level motion displacement between each input image and its corresponding edited target.
We measure motion changes in the image pairs with optical flow, the calculation of which is explained later in Section \ref{sec:4-method}. 

\begin{figure}[h]
\vspace{-0.3em}
 \centering
\includegraphics[width=0.8\linewidth]{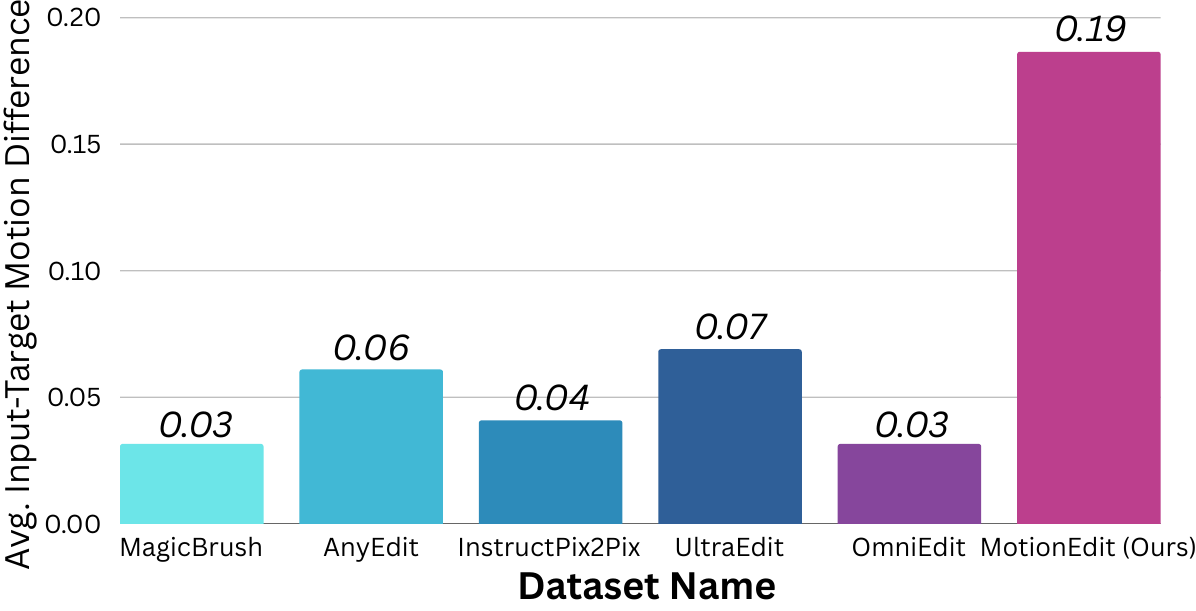}
 \vspace{-0.3em}
 \caption{\label{fig:data-comparison-motion}Comparison of motion difference between before- and post-edit images in different datasets~\citep{Zhang2023MagicBrush,yu2024anyedit,brooks2022instructpix2pix,zhao2024ultraeditinstructionbasedfinegrainedimage,wei2024omniedit}. Our \textsc{MotionEdit} dataset achieves the most significant motion changes.}
\vspace{-0.5em}
\end{figure}
Figure~\ref{fig:data-comparison-motion} reports the average input-target motion magnitude across 6 editing datasets.
Prior datasets such as MagicBrush~\citep{Zhang2023MagicBrush}, AnyEdit~\citep{yu2024anyedit}, InstructPix2Pix~\citep{brooks2022instructpix2pix}, UltraEdit~\citep{zhao2024ultraeditinstructionbasedfinegrainedimage}, and OmniEdit~\citep{wei2024omniedit} contain relatively modest motion changes (typically around $0.05$), whereas our \textsc{MotionEdit} dataset exhibits substantially larger motion differences ($0.19$), representing \textbf{5.8$\times$} greater motion than MagicBrush and OmniEdit and \textbf{3$\times$} that of UltraEdit.
This highlights our contribution of a significantly more challenging motion editing dataset with substantial motion transformation.
\section{Learning Motion Image Editing}
\label{sec:4-method}

\begin{table*}[!t]
\small
\centering
\scalebox{0.9}{
\begin{tabular}{l|cccc|c|c}
\toprule
\multirow{3}*{\textbf{Model}} &
\multicolumn{6}{c}{\textbf{MotionEdit-Bench}}  \\
\cmidrule(lr){2-7}
  & \textbf{Overall$\uparrow$}  
  & \textbf{Fidelity$\uparrow$} 
 & \textbf{Preservation$\uparrow$} 
 & \textbf{Coherence$\uparrow$}  
 & \textbf{Motion Alignment Score (MAS)$\uparrow$} 
 & \textbf{Win Rate$\uparrow$} \\ 
\midrule
Instruct-P2P~\cite{brooks2022instructpix2pix} & 1.30 & 1.32 & 1.29 & 1.29 & 34.15 & 16.09 \\
AnyEdit~\cite{yu2024anyedit} & 1.31 & 1.32 & 1.32 & 1.30 & 35.11 & 16.88 \\
MagicBrush~\cite{Zhang2023MagicBrush} & 1.50 & 1.58 & 1.47 & 1.44 &  44.24 & 19.51 \\
UltraEdit~\cite{zhao2024ultraeditinstructionbasedfinegrainedimage} & 2.42 & 1.88 & 2.09 & 2.13 & 47.18 & 28.33 \\
UniWorld-V1~\citep{lin2025uniworld} & 2.87 & 2.96 & 2.76 & 2.88 & 55.37 & 41.14  \\
Step1X-Edit~\cite{liu2025step1x-edit} & 4.02 & 4.04 & 3.99 & 4.02 & 52.98 & 61.14 \\
BAGEL~\cite{deng2025bagel} & 4.10 & 4.24 & 4.01 & 4.06 & 51.83 & 61.46  \\
\midrule
FLUX.1 Kontext [Dev]~\cite{labs2025flux1kontextflowmatching} & 3.84 & 3.89 & 3.79 & 3.83  & 53.73 & 57.71 \\
\rowcolor{blue!14} \textit{\;\; \; +\textsc{\textbf{MotionNFT}} (Ours)} & \textbf{4.25} & \textbf{4.33} & \textbf{4.16} & \textbf{4.25}  & \textbf{55.45}  &  \textbf{64.95} \\
\midrule
Qwen-Image-Edit~\cite{wu2025qwenimagetechnicalreport} & 4.65 & 4.70 & 4.59 & 4.66  & 56.46  & 72.80 \\
\rowcolor{blue!14}  \textit{\;\; \; +\textsc{\textbf{MotionNFT}} (Ours)}  & \textbf{4.72}  & \textbf{4.79} & \textbf{4.63} & \textbf{4.74}  & \textbf{57.23} & \textbf{73.67} \\
\bottomrule
\end{tabular}}
% \vspace{-0.6em}
\caption{\label{tab:edit_comparison} Quantitative results on \textsc{MotionEdit-Bench}. 
Among existing methods, Step1X-Edit and BAGEL achieve the strongest motion-editing performance, while diffusion-based editors such as AnyEdit and MagicBrush perform poorly across both generative and discriminative metrics. 
FLUX.1~Kontext and Qwen-Image-Edit models trained with MotionNFT yields the best overall results: for both models, applying MotionNFT boosts all generative metrics, MAS and pairwise win rate.
}
\vspace{-0.2em}
\end{table*}

\subsection{Preliminaries}
\noindent\textbf{Flow Matching Models} 
Recent progress in diffusion models has shifted from Denoising Diffusion Probabilistic Models (DDPMs)~\citep{Rombach_2022_CVPR} to Flow Matching Models (FMMs)~\citep{labs2025flux1kontextflowmatching}.
Given noisy sample $z_t$ and conditioning $c$, FMMs reformulate the noise prediction process in DDPMS by estimates a deterministic \emph{velocity field} $v$ that transports $z_t$ toward its clean counterpart. 
As a result, inference for FMMs reduces to the ODE $dz_t = v_\theta(z_t, t, c)\,dt$, 
which enables efficient generation compared to DDPM sampling.

\noindent\textbf{Diffusion Negative-aware Finetuning (NFT)}
DiffusionNFT~\citep{zheng2025diffusionnft} enhances FMM reward training by learning not only a \emph{positive} velocity $v^{+}(x_t, c, t)$ that the model should move toward, but also a \emph{negative} velocity $v^{-}(x_t, c, t)$ that it should avoid.  
The training objective is:
\vspace{-0.4em}
\begin{equation}
\label{eq:diffusion-nft}
\small
\begin{split}
\mathcal{L}(\theta)
=
&\mathbb{E}_{c,\;\pi^{\mathrm{old}}(x_0 \mid c),\;t}
\Big[ r \,\| v_\theta^{+}(x_t, c, t) - v \|_2^2
\\
&\quad\; + (1-r)\, \| v_\theta^{-}(x_t, c, t) - v \|_2^2
\Big],
\end{split}
\vspace{-0.4em}
\end{equation}
where $v$ is the target velocity and $v_\theta^{+}, v_\theta^{-}$ are defined as linear combinations of the old and current policies:
\vspace{-0.4em}
\begin{equation}
\small
\begin{aligned}
v_\theta^{+}(x_t,c,t)
&= (1-\beta)\,v^{\mathrm{old}}(x_t,c,t)
   + \beta\,v_\theta(x_t,c,t), \\
v_\theta^{-}(x_t,c,t)
&= (1+\beta)\,v^{\mathrm{old}}(x_t,c,t)
   - \beta\,v_\theta(x_t,c,t).
\end{aligned}
\vspace{-0.3em}
\end{equation}
A key challenge is obtaining a calibrated reward $r$ that accurately reflects whether a sample should be treated as ``positive''.  
Since raw rewards may differ in scale or distribution, DiffusionNFT transforms them into an \emph{optimality reward}:
\begin{equation}
\scriptsize
r(x_0,c)
=
\frac{1}{2}
+
\frac{1}{2}
\operatorname{clip}
\!\left[
\frac{
r^{\mathrm{raw}}(x_0,c)
-
\mathbb{E}_{\pi^{\mathrm{old}}(\cdot \mid c)}
\big[r^{\mathrm{raw}}(x_0,c)\big]
}{
Z_c
},
-1, 1
\right],
\end{equation}
where $Z_c$ is a normalization factor (e.g., the global reward standard deviation).  
This normalization stabilizes learning and ensures consistent positive/negative assignment across prompts and reward models.

\subsection{MotionNFT: Motion-Aware Reward for NFT}
We introduce \textbf{MotionNFT}, a motion-aware reward framework designed for NFT training on motion-editing tasks. Since our objective is to evaluate how accurately a model applies the intended action to subjects and objects, our reward function must quantify the alignment between model-predicted motion and the ground-truth motion edit. Inspired by the use of optical flow for measuring motion between consecutive video frames, we adopt an optical-flow–based \textbf{motion-centric scoring framework} that treats each input–edit pair as an implicit ``before–after’’ sequence.

Given a triplet $\mathbf{X} = (\mathbf{I}_{\text{orig}}, \mathbf{I}_{\text{edited}}, \mathbf{I}_{\text{gt}})$, we compute optical flow fields using a pretrained estimator~\citep{xu2023unifying}. The predicted motion is $\mathbf{V}_{\text{pred}} = \mathcal{F}(\mathbf{I}_{\text{orig}}, \mathbf{I}_{\text{edited}})$ and the ground-truth motion is $\mathbf{V}_{\text{gt}} = \mathcal{F}(\mathbf{I}_{\text{orig}}, \mathbf{I}_{\text{gt}})$, where each flow lies in $\mathbb{R}^{H \times W \times 2}$. We normalize both flows by the image diagonal to ensure scale consistency across resolutions.

\textbf{Motion magnitude consistency.}  
We measure the deviation between flow magnitudes using a robust $\ell_{1}$ distance:  
$\mathcal{D}_{\text{mag}} = \tfrac{1}{HW}\sum_{i,j}(\|\tilde{\mathbf{V}}_{\text{pred}}(i,j)-\tilde{\mathbf{V}}_{\text{gt}}(i,j)\|_{1}+\varepsilon)^{q}$,  
where $q\in(0,1)$ is a constant term to suppress outliers.

\textbf{Motion direction consistency.}  
We compute cosine-based directional error between the unit flow vectors  
$e_{\text{dir}}(i,j)=\tfrac{1}{2}\bigl(1-\hat{\mathbf{v}}_{\text{pred}}(i,j)^\top\hat{\mathbf{v}}_{\text{gt}}(i,j)\bigr)$,  
and weight each pixel by its relative ground-truth motion magnitude. The directional misalignment is  
$\mathcal{D}_{\text{dir}} = \tfrac{\sum_{i,j} w(i,j) e_{\text{dir}}(i,j)}{\sum_{i,j} w(i,j)+\varepsilon}$.

\textbf{Movement regularization.}  
To prevent trivial edits that make almost no motion, we compare the average predicted and ground-truth magnitudes:  
$M_{\text{move}}=\max\{0,\ \tau+\tfrac{1}{2}\bar{m}_{\text{gt}}-\bar{m}_{\text{pred}}\}$,  
where $\tau$ is a small positive margin and $\bar{m}$ denotes the spatial mean.

\begin{figure}[!t]
\centering
% \vspace{-4mm}
\includegraphics[width=0.95\linewidth]{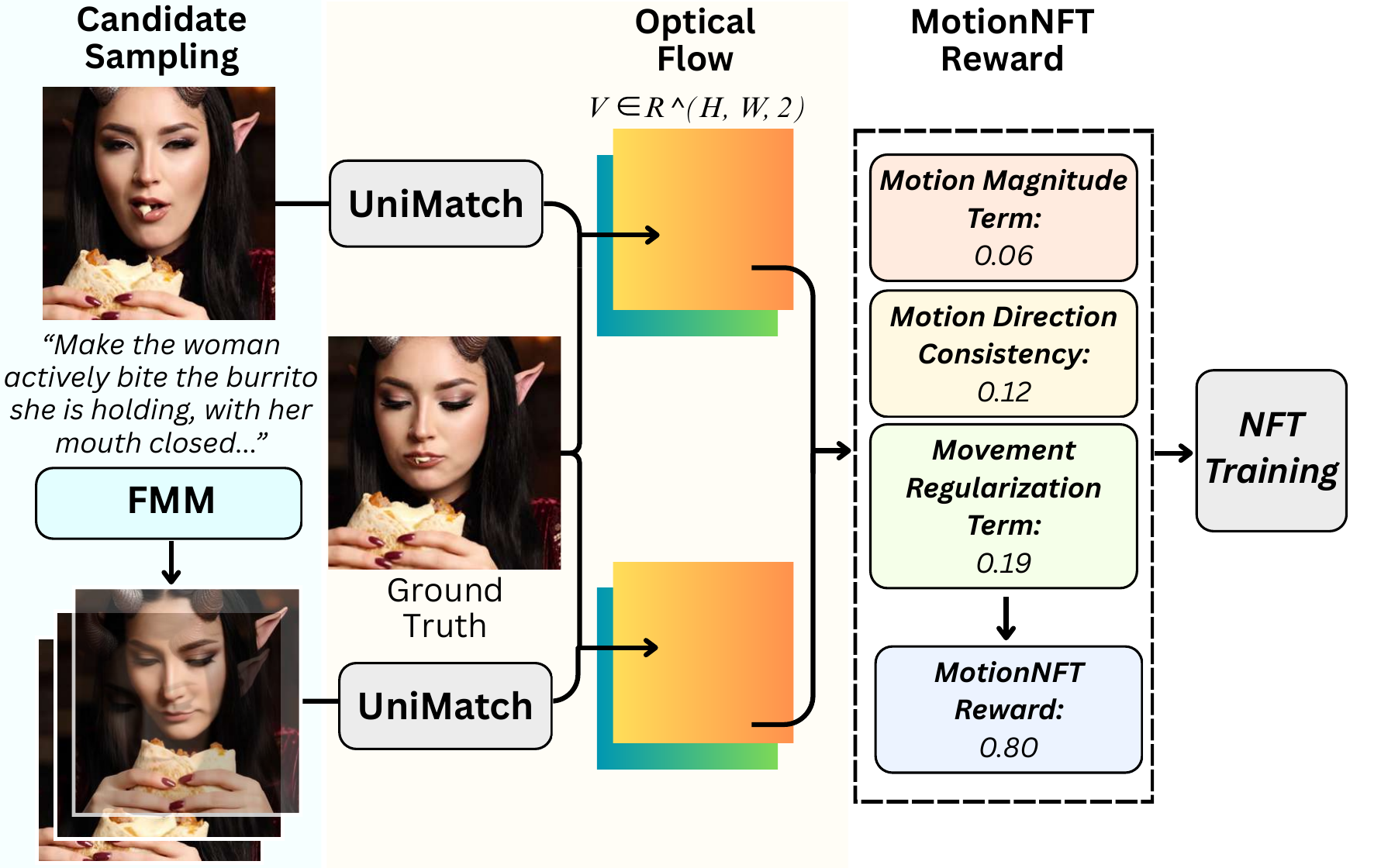}
% \vspace{-1em}
\caption{\label{fig:motionnft-pipeline}MotionNFT's Reward Scoring pipeline. For each sampled model-edited image, we measure the alignment between the input-generated optical flow and the input-ground truth optical flow, obtaining the final reward score.}
\vspace{-0.7em}
\end{figure}

\textbf{Combined reward.}  
We aggregate the three terms into a composite distance  
$\mathcal{D}_{\text{comb}} = \alpha\,\mathcal{D}_{\text{mag}} + \beta\,\mathcal{D}_{\text{dir}} + \lambda_{\text{move}} M_{\text{move}}$  
where $\alpha$, $\beta$, and $\lambda_{\text{move}}$ are constants that balances term scales and weightings.  
The composite distance is then normalized and clipped:  
$\tilde{D} = \mathrm{clip}\;\big((\mathcal{D}_{\text{comb}}-\mathcal{D}_{\min}^*)/(\mathcal{D}_{\max}-\mathcal{D}_{\min}^*), 0, 1\big)$,  
and converted into a continuous reward $r_{\text{cont}} = 1 - \tilde{D}$.  
Finally, we quantize it into 6 discrete reward levels:  
$r_{\text{motion}} = \tfrac{1}{5}\,\mathrm{round}(5\, r_{\text{cont}}) \in \{0.0,\, 0.2,\, 0.4,\, 0.6,\, 0.8,\, 1.0\} $.
The resulting scalar reward is used to compute optimality rewards and update the policy model $v_{\theta}$ under the DiffusionNFT objective (Eq.~\ref{eq:diffusion-nft}). Figure~\ref{fig:motionnft-pipeline} illustrates the MotionNFT reward pipeline.

\vspace{-0.7em}

\section{Experiments}

\begin{figure*}[t]
 \centering
 % \vspace{-4mm}
 \includegraphics[width=\linewidth]{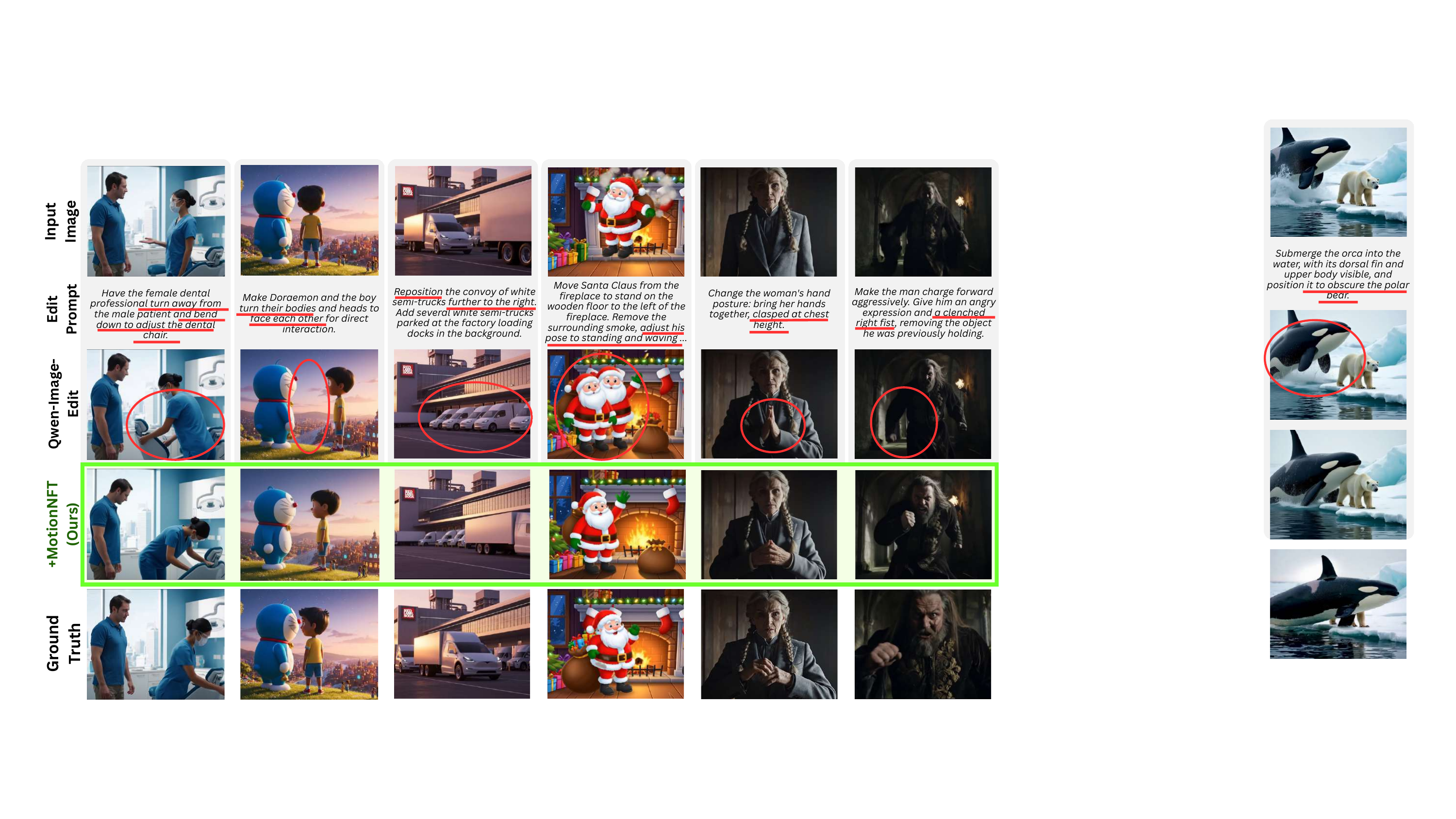}
 \vspace{-1em}
 \caption{Qualitative examples of our MotionNFT. The baseline \textsc{Qwen-Image-Edit}~\citep{wu2025qwenimagetechnicalreport} model often fails to execute the instructed motion (circled regions), producing edits that do not match the required action change (red underlines). With MotionNFT training, the model succeeds in performing precise motion edits that closely align with the ground-truth transformations.}
 % \lk{include some additional text to describe our advantage cases.}}
 \vspace{-0.3em}
 \label{fig:motionnft-results}
% \vspace{-0.6em}
\end{figure*}

\subsection{Experimental Setup}
We provide important details of our experimental setups.
Full details are reported in the \textit{Additional Experiment Details} Appendix section.

\myparagraph{MotionNFT Training}
We use \textsc{FLUX.1 Kontext [Dev]}~\citep{labs2025flux1kontextflowmatching} and \textsc{Qwen-Image-Edit}~\citep{wu2025qwenimagetechnicalreport} as base models for MotionNFT training. 
% For each model, we allocate 2 nodes, each with 8 NVIDIA H20 GPUs.
Following \citet{lin2025uniworld}'s implementation, we use Fully Sharded Data Parallelism (FSDP) for text encoder and apply gradient checkpointing in training for GPU memory usage optimization.
To improve models' motion image editing capabilities while preserving general image editing ability, we employ a multi-score reward formulation with a weighted combination of (\emph{i}) 50\% our optical flow-based \emph{Motion Reward} $r_motion$ and (\emph{ii}) 50\% MLLM reward proposed by \citet{lin2025uniworld}. 
For MLLM-based evaluation, we serve a \textsc{Qwen2.5-VL-32B-Instruct}~\citep{Qwen2.5-VL} model via vLLM on a separate node that performs online scoring throughout training.
The optical flow component of our reward leverages a lightweight UniMatch model (335.6M parameters), which we run directly on the training nodes to provide efficient motion-level guidance.

\myparagraph{Benchmarked Image Editing Models}
We evaluate 9 open-source models on \textsc{MotionEdit-Bench}: Instruct-P2P~\citep{brooks2022instructpix2pix}, MagicBrush~\citep{Zhang2023MagicBrush}, AnyEdit~\citep{yu2024anyedit}, UltraEdit~\citep{zhao2024ultraeditinstructionbasedfinegrainedimage}, Step1X-Edit~\citep{liu2025step1x-edit}, BAGEL~\citep{deng2025bagel}, UniWorld-V1~\citep{lin2025uniworld}, FLUX.1 Kontext [Dev]~\citep{labs2025flux1kontextflowmatching}, and Qwen-Image-Edit~\citep{wu2025qwenimagetechnicalreport}.

\subsection{Evaluation Metrics}

\myparagraph{Generative Metrics.}
Following~\citet{luo2025editscore} and~\citet{lin2025uniworld}, we use an MLLM to evaluate edited images with four generative metrics: \textit{Fidelity}, \textit{Preservation}, \textit{Coherence}, and their \textit{Overall} average.
We choose to use Google's Gemini~\citep{team2023gemini} as the MLLM evaluator and use evaluation prompts adapted from the “action’’ category of~\citet{luo2025editscore}.

\myparagraph{Discriminative Motion Alignment Score (MAS).}
% Since motion editing requires verifying whether a model applies the correct action to subjects and objects, 
To complement the MLLM generative metric with deterministic assessment, we introduce an optical flow–based Motion Alignment Score (MAS) to measure how well the model understands and performs the correct motion change in images.
% well a model make changes to reflect the intended motion of subjects and objects.
MAS combines the \emph{motion magnitude consistency} term $\mathcal{D}_{\text{mag}}$ and the \emph{motion direction consistency} term $\mathcal{D}_{\text{dir}}$ from Section \ref{sec:4-method} into a single motion alignment metric:
% We first form a composite distance combining magnitude and direction consistency, 
$\mathcal{D}_{\mathrm{ovl}} = \alpha\,\mathcal{D}_{\mathrm{mag}} + (1-\alpha)\,\mathcal{D}_{\mathrm{dir}}$, where $\alpha$ is a constant term balancing scales.
Then, we normalize $\mathcal{D}_{\mathrm{ovl}}$ and convert it into: $\mathrm{MAS} = 100.00 \cdot (1 - \mathrm{clip}\;\big((\mathcal{D}_{\mathrm{ovl}} - d_{\min})/(d_{\max} - d_{\min}),\,0,1\big))$. Higher scores indicate closer alignment. If the predicted motion is nearly static compared to ground truth, i.e., $\mathbb{E}[m_{\mathrm{pred}}]/\mathbb{E}[m_{\mathrm{gt}}] < \rho_{\min}$, we assign $\mathrm{MAS}=0$.

\subsection{Quantitative Evaluation Results}
Table~\ref{tab:edit_comparison} reports quantitative performance of 9 image editing models on \textsc{MotionEdit-Bench}.
The first 4 columns shows MLLM generative ratings on a 0–5 scale. 
Our optical flow-based \emph{MAS} metric measures motion consistency on a 0–100 scale. 
The \emph{Win Rate} reflects the percentage of pairwise comparisons in which a model’s output received a higher average MLLM score than a competing one.

\myparagraph{\#1: Improved Motion Editing Quality.} Across both base models, MotionNFT consistently improves all aspects of generation quality on motion editing, as measured by the generative evaluator. 
When applied to \textsc{FLUX.1 Kontext}, MotionNFT raises the Overall score from 3.84 to 4.25 ($+10.68\%$), with notable gains in Fidelity ($+0.44$) and Coherence ($+0.42$). 
For \textsc{Qwen-Image-Edit}, MotionNFT also improves the Overall score from $4.65$ to $4.72$.

\myparagraph{\#2: Enhanced Motion Alignment.} 
MotionNFT yields substantial improvements in MAS, highlighting its effectiveness in producing motion changes consistent with the ground-truth edits. 
On \textsc{FLUX.1 Kontext}, MotionNFT increases MAS from 53.73 to 55.45, while on \textsc{Qwen-Image-Edit}, MAS improves from 56.46 to 57.23. 
These gains are achieved despite the strong baselines and show that our flow-based reward provides meaningful guidance for learning spatial and motion-aware transformations.

\myparagraph{\#3: Strong Pairwise Preference Performance.}
MotionNFT also achieves higher win rates relative to all evaluated models.
% reflecting improved perceptual quality under direct comparison settings. 
For \textsc{FLUX.1 Kontext}, MotionNFT boosts win rate from 57.97\% to 65.16\% ($+12.40\%$), and from 72.99\% to 73.87\% for \textsc{Qwen-Image-Edit}. 
These results show that MotionNFT produces more accurate motion edits that are more frequently preferred over outputs of other models.

\subsection{Qualitative Evaluation Results}
% We analyze the results along three core dimensions: Fidelity (correctness of the motion change), Preservation (identity and appearance consistency), and Coherence (physical plausibility and spatial alignment). 
Figure~\ref{fig:model-comparison} and Figure~\ref{fig:motionnft-results} illustrate representative qualitative results on \textsc{MotionEdit}. 

% Across a diverse range of prompts, e
\myparagraph{\# 1: Existing models struggle to perform correct motion edits.}
We observe that even state-of-the-art open-sourced image editing models like FLUX.1 Kontext and Qwen-Image-Edit struggle to correctly perform motion-centric changes like turning body directions. 
% As shown in the first row of Figure~\ref{fig:model-comparison}, t
These models often preserve the original pose or only apply superficial appearance changes. 
This highlights the crucial bottleneck in translating motion-related language instructions into coherent image subject / object transformations.

\myparagraph{\#2: \textsc{MotionNFT} improves motion editing capability.}
Training with \textsc{MotionNFT} enables Qwen-Image-Edit to produce outputs that more faithfully reflect the intended motion, e.g. rotating character directions, adjusts limb and torso positions to reflect bending or turning actions.
% As demonstrated in Figure~\ref{fig:qualitative}, 
Additionally, the resulting edits preserve identity and scene context while achieving the targeted motion change, closely matching the ground-truth transformations. 
These observations validates the effectiveness of incorporating motion-centric guidance in MotionNFT to execute meaningful, structure-aware motion edits that current image editing models consistently fall short in achieving.

\subsection{Ablation Studies}

\myparagraph{General Image Editing Performance}
To verify that MotionNFT preserves a model’s general editing ability, we follow previous work~\citep{lin2025uniworld} and conduct evaluation on ImgEdit-Bench~\citep{ye2025imgedit}, a comprehensive benchmark covering 8 editing subtasks. 
Table~\ref{tab:edit_comparison_imgedit} shows that MotionNFT consistently improves or maintains performance across all categories for both \textsc{FLUX.1 Kontext} and \textsc{Qwen-Image-Edit}, even yielding higher overall scores. 
Results confirm that MotionNFT can enhance models' motion editing performance without trading off general editing quality.

\begin{table}[h]
% \vspace{-0.3em}
\scriptsize
\centering
% 1 (Model) | 4 (MP) | 7 (ImgEdit) | 1 (Overall)  ==> 13 columns total
\begin{tabular}{p{0.11\textwidth}|p{0.0118\textwidth}p{0.0118\textwidth}p{0.0118\textwidth}p{0.0118\textwidth}p{0.0118\textwidth}p{0.0118\textwidth}p{0.0118\textwidth}p{0.015\textwidth}|p{0.0135\textwidth}}
\toprule
\textbf{Model} &
\multicolumn{8}{c}{\textbf{ImgEdit-Bench}}  \\
\cmidrule(lr){2-10}
 & \textbf{Add} & \textbf{Adj.} & \textbf{Rpl.} & \textbf{Rem.} & \textbf{Bck.} & \textbf{Stl.} & \textbf{Hyb.}  & \textbf{Act.} 
 & \textbf{Ovl.$\uparrow$}\\ % Overall column has its header already
\midrule
FLUX.1 Kontext & 3.54 & 2.90 & 3.73 & 2.89 &  3.59  & 3.96  & 2.90 & 2.56  &  3.26 \\
\rowcolor{blue!14} \textit{\; \textbf{+ \textsc{MotionNFT}}} & \textbf{3.71} & \textbf{3.28} & \textbf{3.93} & \textbf{3.05} & \textbf{3.72} & \textbf{4.41}  & \textbf{2.99} & \textbf{2.85} & \textbf{3.50} \\ 
\midrule
Qwen-Image-Edit  & 4.20 & 3.70 & 4.22 & 4.20 & 4.17 & 4.60 & 3.55 & \textbf{4.03} & 4.08 \\
\rowcolor{blue!14} \textit{\; \textbf{+ \textsc{MotionNFT}}}   & \textbf{4.31} & \textbf{3.72} & \textbf{4.46} & \textbf{4.30} & \textbf{4.21} & \textbf{4.67} & \textbf{3.96} & 3.87 & \textbf{4.20} \\ % flow score 3 + mllm, step 210
% \midrule
\bottomrule
\end{tabular}
\vspace{-0.2em}
\caption{\label{tab:edit_comparison_imgedit} Results on ImgEdit-Bench~\citep{ye2025imgedit} MotionNFT not only preserves, but oftentimes boosts general editing performances.}
% \vspace{-0.5em}
\end{table}

\myparagraph{Comparison with MLLM-only Reward}
To verify the effect of MotionNFT's supervision, we compare MotionNFT against the MLLM-only RL framework in UniWorld-V2~\citep{lin2025uniworld}. 
Table~\ref{tab:comparison-uniworld} shows that while MLLM-only training yields modest improvements over the base models, MotionNFT consistently achieves higher overall edit quality, better motion alignment, and superior win rates across both base models. 
These results demonstrate that incorporating optical flow-based motion guidance yields more targeted and effective motion-editing capabilities.
% MotionNFT outperforms previously proposed MLLM-only RL training framework in Uniworld-V2~\citep{lin2025uniworld} on motion editing task.

\begin{table}[h]
% \vspace{-0.3em}
\small
\centering
\resizebox{0.45\textwidth}{!}{
% 1 (Model) | 4 (MP) | 7 (ImgEdit) | 1 (Overall)  ==> 13 columns total
\begin{tabular}{l|p{0.08\textwidth}p{0.08\textwidth}p{0.1\textwidth}}
\toprule
\multirow{2}*{\textbf{Model}} &
\multicolumn{3}{c}{\textbf{MotionEdit-Bench}}  \\
\cmidrule(lr){2-4}
  & \textbf{Overall. $\uparrow$} 
 & \textbf{MAS $\uparrow$} 
 & \textbf{Win Rate $\uparrow$} \\ 
\midrule
FLUX.1 Kontext  & 3.84  & 53.73 & 57.71 \\
\textit{\;\;\; + UniWorld-V2\cite{lin2025uniworld}} & 4.20    & 54.58 & 63.76 \\
\rowcolor{blue!14} \textit{\;\; \; +\textsc{\textbf{MotionNFT}} (Ours)} & \textbf{4.25}  & \textbf{55.45}  &  \textbf{64.95} \\
\midrule
Qwen-Image-Edit & 4.65   & 56.46  & 72.80 \\
\textit{\;\;\; + UniWorld-V2\cite{lin2025uniworld}}   & 4.70  & 56.46 & 72.56 \\
% \textit{\;\;\; + UniWorld-V2\cite{lin2025uniworld}} & Rerun & 4.80 & 4.57 & 4.73   &  & \\
\rowcolor{blue!14}  \textit{\;\; \; +\textsc{\textbf{MotionNFT}} (Ours)}  & \textbf{4.72}  & \textbf{57.23} & \textbf{73.67} \\
\bottomrule
\end{tabular}}
\vspace{-0.2em}
\caption{\label{tab:comparison-uniworld}Comparison to training with MLLM-based reward~\citep{lin2025uniworld} only. Incorporating MotionNFT yields noticeable improvements MLLM-scored Overall editing quality, optical flow-based Motion Alignment Score, and the pairwise Win Rate across all models.
}
\vspace{-0.8em}
\end{table}
\vspace{-0.3em}
\section{Conclusion}
We introduced \textsc{MotionEdit}, a high-quality dataset and benchmark for the novel motion image editing task, aiming at correct modifying subject actions and interactions in images while preserving identity and scene consistency.
% By mining coherent motion transitions from continuous video frames, our scalable pipeline produces diverse, faithful input–edit pairs that fill a critical gap left by existing appearance-centric editing datasets.
To improve model performance on this challenging task, we proposed \textsc{MotionNFT}, a motion-guided negative-aware finetuning framework that integrates an optical-flow–based motion reward for training. 
MotionNFT provides supervision on motion magnitude and direction, enabling models to understand and perform motion transformations that existing models consistently struggle with.
Both quantitative and qualitative experiment results demonstrate that MotionNFT delivers consistent gains across generative quality, motion alignment, and preference win rate on two strong base models, FLUX.1 Kontext and Qwen-Image-Edit.

% \begin{itemize}
%     \item Summary: Reiterate your three main achievements:
%     \begin{itemize}
%         \item Defining the new Moving Pixels task.
%         \item Creating the comprehensive benchmark and training data (Bench and Train).
%         \item Proposing a state-of-the-art motion-aware editing framework.
%     \end{itemize}
%     \item Impact: Emphasize the significant gains achieved on motion metrics and perceptual quality.
%     \item Future Work: Mention the open-source release of the code, dataset, and evaluation toolkit to standardize assessment and spur future research.
% \end{itemize}

{
    \small
    \bibliographystyle{ieeenat_fullname}
    \bibliography{main}

@String(CVPR= {IEEE Conf. Comput. Vis. Pattern Recog.})

@String(ICCV= {Int. Conf. Comput. Vis.})

@String(CVPR  = {CVPR})

@String(ICCV  = {ICCV})

@misc{labs2025flux1kontextflowmatching,
      title={FLUX.1 Kontext: Flow Matching for In-Context Image Generation and Editing in Latent Space},
      author={Black Forest Labs and Stephen Batifol and Andreas Blattmann and Frederic Boesel and Saksham Consul and Cyril Diagne and Tim Dockhorn and Jack English and Zion English and Patrick Esser and Sumith Kulal and Kyle Lacey and Yam Levi and Cheng Li and Dominik Lorenz and Jonas Müller and Dustin Podell and Robin Rombach and Harry Saini and Axel Sauer and Luke Smith},
      year={2025},
      eprint={2506.15742},
      archivePrefix={arXiv},
      primaryClass={cs.GR},
      url={https://arxiv.org/abs/2506.15742},
}

@misc{wu2025qwenimagetechnicalreport,
      title={Qwen-Image Technical Report}, 
      author={Chenfei Wu and Jiahao Li and Jingren Zhou and Junyang Lin and Kaiyuan Gao and Kun Yan and Sheng-ming Yin and Shuai Bai and Xiao Xu and Yilei Chen and Yuxiang Chen and Zecheng Tang and Zekai Zhang and Zhengyi Wang and An Yang and Bowen Yu and Chen Cheng and Dayiheng Liu and Deqing Li and Hang Zhang and Hao Meng and Hu Wei and Jingyuan Ni and Kai Chen and Kuan Cao and Liang Peng and Lin Qu and Minggang Wu and Peng Wang and Shuting Yu and Tingkun Wen and Wensen Feng and Xiaoxiao Xu and Yi Wang and Yichang Zhang and Yongqiang Zhu and Yujia Wu and Yuxuan Cai and Zenan Liu},
      year={2025},
      eprint={2508.02324},
      archivePrefix={arXiv},
      primaryClass={cs.CV},
      url={https://arxiv.org/abs/2508.02324}, 
}

@article{deng2025bagel,
  title   = {Emerging Properties in Unified Multimodal Pretraining},
  author  = {Deng, Chaorui and Zhu, Deyao and Li, Kunchang and Gou, Chenhui and Li, Feng and Wang, Zeyu and Zhong, Shu and Yu, Weihao and Nie, Xiaonan and Song, Ziang and Shi, Guang and Fan, Haoqi},
  journal = {arXiv preprint arXiv:2505.14683},
  year    = {2025}
}

@misc{openai2025imagegen,
  author = {OpenAI},
  title = {Image Generation API},
  year = {2025},
  note = {\url{https://openai.com/index/image-generation-api/}}
}

@misc{nanobanana,
  author = {Google},
  title = {Gemini Image Generation API},
  year = {2025},
  note = {\url{https://ai.google.dev/gemini-api/docs/image-generation/}}
}

@article{seedream2025seedream,
  title={Seedream 4.0: Toward next-generation multimodal image generation},
  author={Seedream, Team and Chen, Yunpeng and Gao, Yu and Gong, Lixue and Guo, Meng and Guo, Qiushan and Guo, Zhiyao and Hou, Xiaoxia and Huang, Weilin and Huang, Yixuan and others},
  journal={arXiv preprint arXiv:2509.20427},
  year={2025}
}

@article{cao2025hunyuanimage,
  title={Hunyuanimage 3.0 technical report},
  author={Cao, Siyu and Chen, Hangting and Chen, Peng and Cheng, Yiji and Cui, Yutao and Deng, Xinchi and Dong, Ying and Gong, Kipper and Gu, Tianpeng and Gu, Xiusen and others},
  journal={arXiv preprint arXiv:2509.23951},
  year={2025}
}

@article{lin2025uniworld,
  title={UniWorld: High-Resolution Semantic Encoders for Unified Visual Understanding and Generation},
  author={Lin, Bin and Li, Zongjian and Cheng, Xinhua and Niu, Yuwei and Ye, Yang and He, Xianyi and Yuan, Shenghai and Yu, Wangbo and Wang, Shaodong and Ge, Yunyang and others},
  journal={arXiv preprint arXiv:2506.03147},
  year={2025}
}

@article{brooks2022instructpix2pix,
  title={InstructPix2Pix: Learning to Follow Image Editing Instructions},
  author={Brooks, Tim and Holynski, Aleksander and Efros, Alexei A},
  journal={arXiv preprint arXiv:2211.09800},
  year={2022}
}

@misc{zhao2024ultraeditinstructionbasedfinegrainedimage,
      title={UltraEdit: Instruction-based Fine-Grained Image Editing at Scale}, 
      author={Haozhe Zhao and Xiaojian Ma and Liang Chen and Shuzheng Si and Rujie Wu and Kaikai An and Peiyu Yu and Minjia Zhang and Qing Li and Baobao Chang},
      year={2024},
      eprint={2407.05282},
      archivePrefix={arXiv},
      primaryClass={cs.CV},
      url={https://arxiv.org/abs/2407.05282}, 
}

@article{wei2024omniedit,
  title={OmniEdit: Building Image Editing Generalist Models Through Specialist Supervision},
  author={Wei, Cong and Xiong, Zheyang and Ren, Weiming and Du, Xinrun and Zhang, Ge and Chen, Wenhu},
  journal={arXiv preprint arXiv:2411.07199},
  year={2024}
}

@inproceedings{Zhang2023MagicBrush,
        title={MagicBrush: A Manually Annotated Dataset for Instruction-Guided Image Editing},
        author={Kai Zhang and Lingbo Mo and Wenhu Chen and Huan Sun and Yu Su},
        booktitle={Advances in Neural Information Processing Systems},
        year={2023}
}

@article{zheng2025diffusionnft,
  title={DiffusionNFT: Online Diffusion Reinforcement with Forward Process},
  author={Zheng, Kaiwen and Chen, Huayu and Ye, Haotian and Wang, Haoxiang and Zhang, Qinsheng and Jiang, Kai and Su, Hang and Ermon, Stefano and Zhu, Jun and Liu, Ming-Yu},
  journal={arXiv preprint arXiv:2509.16117},
  year={2025}
}

@inproceedings{mengsdedit,
  title={SDEdit: Guided Image Synthesis and Editing with Stochastic Differential Equations},
  author={Meng, Chenlin and He, Yutong and Song, Yang and Song, Jiaming and Wu, Jiajun and Zhu, Jun-Yan and Ermon, Stefano},
  booktitle={International Conference on Learning Representations}
}

@article{lin2024schedule,
  title={Schedule your edit: A simple yet effective diffusion noise schedule for image editing},
  author={Lin, Haonan and Chen, Yan and Wang, Jiahao and An, Wenbin and Wang, Mengmeng and Tian, Feng and Liu, Yong and Dai, Guang and Wang, Jingdong and Wang, Qianying},
  journal={Advances in Neural Information Processing Systems},
  volume={37},
  pages={115712--115756},
  year={2024}
}

@article{yu2024anyedit,
  title={AnyEdit: Mastering Unified High-Quality Image Editing for Any Idea},
  author={Yu, Qifan and Chow, Wei and Yue, Zhongqi and Pan, Kaihang and Wu, Yang and Wan, Xiaoyang and Li, Juncheng and Tang, Siliang and Zhang, Hanwang and Zhuang, Yueting},
  journal={arXiv preprint arXiv:2411.15738},
  year={2024}
}

@inproceedings{sun2018pwc,
  title={Pwc-net: Cnns for optical flow using pyramid, warping, and cost volume},
  author={Sun, Deqing and Yang, Xiaodong and Liu, Ming-Yu and Kautz, Jan},
  booktitle={Proceedings of the IEEE conference on computer vision and pattern recognition},
  pages={8934--8943},
  year={2018}
}

@inproceedings{teed2020raft,
  title={Raft: Recurrent all-pairs field transforms for optical flow},
  author={Teed, Zachary and Deng, Jia},
  booktitle={European conference on computer vision},
  pages={402--419},
  year={2020},
  organization={Springer}
}

@inproceedings{xu2022gmflow,
  title={GMFlow: Learning Optical Flow via Global Matching},
  author={Xu, Haofei and Zhang, Jing and Cai, Jianfei and Rezatofighi, Hamid and Tao, Dacheng},
  booktitle={Proceedings of the IEEE/CVF Conference on Computer Vision and Pattern Recognition},
  pages={8121-8130},
  year={2022}
}

@inproceedings{jiang2021learning,
  title={Learning to estimate hidden motions with global motion aggregation},
  author={Jiang, Shihao and Campbell, Dylan and Lu, Yao and Li, Hongdong and Hartley, Richard},
  booktitle={Proceedings of the IEEE/CVF international conference on computer vision},
  pages={9772--9781},
  year={2021}
}

@article{xu2023unifying,
  title={Unifying Flow, Stereo and Depth Estimation},
  author={Xu, Haofei and Zhang, Jing and Cai, Jianfei and Rezatofighi, Hamid and Yu, Fisher and Tao, Dacheng and Geiger, Andreas},
  journal={IEEE Transactions on Pattern Analysis and Machine Intelligence},
  year={2023}
}

@inproceedings{haa500,
    title={HAA500: Human-Centric Atomic Action Dataset with Curated Videos},
    author={Jihoon Chung and Cheng-hsin Wuu and Hsuan-ru Yang and Yu-Wing Tai and Chi-Keung Tang},
    booktitle = {ICCV 2021}
}

@article{kay2017kinetics,
  title={The kinetics human action video dataset},
  author={Kay, Will and Carreira, Joao and Simonyan, Karen and Zhang, Brian and Hillier, Chloe and Vijayanarasimhan, Sudheendra and Viola, Fabio and Green, Tim and Back, Trevor and Natsev, Paul and others},
  journal={arXiv preprint arXiv:1705.06950},
  year={2017}
}

@misc{veo3_2025,
  title        = {Veo 3},
  author       = {{Google DeepMind}},
  year         = {2025},
  howpublished = {\url{https://deepmind.google/models/veo/}},
  note         = {Accessed: 2025-11}
}

@misc{klingai_2025,
  title        = {Kling AI},
  author       = {{Kuaishou Technology}},
  year         = {2025},
  howpublished = {\url{https://app.klingai.com/global/image-to-video/}},
  note         = {Accessed: 2025-11}
}

@article{wang2024vidprom,
  title={VidProM: A Million-scale Real Prompt-Gallery Dataset for Text-to-Video Diffusion Models},
  author={Wang, Wenhao and Yang, Yi},
  booktitle={Thirty-eighth Conference on Neural Information Processing Systems},
  year={2024},
  url={https://openreview.net/forum?id=pYNl76onJL}
}

@misc{klingai_dataset_2025,
  title        = {KlingAI Video Dataset},
  author       = {Nyuuzyou},
  year         = {2025},
  publisher    = {Hugging Face},
  howpublished = {\url{https://huggingface.co/datasets/nyuuzyou/klingai}},
  note         = {Accessed: 2025-05}
}

@article{team2023gemini,
  title={Gemini: a family of highly capable multimodal models},
  author={Team, Gemini and Anil, Rohan and Borgeaud, Sebastian and Alayrac, Jean-Baptiste and Yu, Jiahui and Soricut, Radu and Schalkwyk, Johan and Dai, Andrew M and Hauth, Anja and Millican, Katie and others},
  journal={arXiv preprint arXiv:2312.11805},
  year={2023}
}

@inproceedings{henriksson2025finerweb,
  title={FinerWeb-10BT: Refining Web Data with LLM-Based Line-Level Filtering},
  author={Henriksson, Erik and Tarkka, Otto and Ginter, Filip},
  booktitle={Proceedings of the Joint 25th Nordic Conference on Computational Linguistics and 11th Baltic Conference on Human Language Technologies (NoDaLiDa/Baltic-HLT 2025)},
  pages={258--268},
  year={2025}
}

@article{wang2025open,
  title={Open-Qwen2VL: Compute-Efficient Pre-Training of Fully-Open Multimodal LLMs on Academic Resources},
  author={Wang, Weizhi and Tian, Yu and Yang, Linjie and Wang, Heng and Yan, Xifeng},
  journal={arXiv preprint arXiv:2504.00595},
  year={2025}
}

@article{cayir2025refine,
  title={Refine-n-Judge: Curating High-Quality Preference Chains for LLM-Fine-Tuning},
  author={Cayir, Derin and Tao, Renjie and Rungta, Rashi and Sun, Kai and Chen, Sean and Khan, Haidar and Kim, Minseok and Reinspach, Julia and Liu, Yue},
  journal={arXiv preprint arXiv:2508.01543},
  year={2025}
}

@inproceedings{chen2024your,
  title={Your Vision-Language Model Itself Is a Strong Filter: Towards High-Quality Instruction Tuning with Data Selection},
  author={Chen, Ruibo and Wu, Yihan and Chen, Lichang and Liu, Guodong and He, Qi and Xiong, Tianyi and Liu, Chenxi and Guo, Junfeng and Huang, Heng},
  booktitle={Findings of the Association for Computational Linguistics ACL 2024},
  pages={4156--4172},
  year={2024}
}

@article{liu2025step1x-edit,
      title={Step1X-Edit: A Practical Framework for General Image Editing}, 
      author={Shiyu Liu and Yucheng Han and Peng Xing and Fukun Yin and Rui Wang and Wei Cheng and Jiaqi Liao and Yingming Wang and Honghao Fu and Chunrui Han and Guopeng Li and Yuang Peng and Quan Sun and Jingwei Wu and Yan Cai and Zheng Ge and Ranchen Ming and Lei Xia and Xianfang Zeng and Yibo Zhu and Binxing Jiao and Xiangyu Zhang and Gang Yu and Daxin Jiang},
      journal={arXiv preprint arXiv:2504.17761},
      year={2025}
}

@InProceedings{Rombach_2022_CVPR,
    author    = {Rombach, Robin and Blattmann, Andreas and Lorenz, Dominik and Esser, Patrick and Ommer, Bj\"orn},
    title     = {High-Resolution Image Synthesis With Latent Diffusion Models},
    booktitle = {Proceedings of the IEEE/CVF Conference on Computer Vision and Pattern Recognition (CVPR)},
    month     = {June},
    year      = {2022},
    pages     = {10684-10695}
}

@article{luo2025editscore,
  title={EditScore: Unlocking Online RL for Image Editing via High-Fidelity Reward Modeling},
  author={Luo, Xin and Wang, Jiahao and Wu, Chenyuan and Xiao, Shitao and Jiang, Xiyan and Lian, Defu and Zhang, Jiajun and Liu, Dong and others},
  journal={arXiv preprint arXiv:2509.23909},
  year={2025}
}

@inproceedings{ren2024diffusion,
  title={Diffusion Policy Policy Optimization},
  author={Ren, Allen Z and Lidard, Justin and Ankile, Lars Lien and Simeonov, Anthony and Agrawal, Pulkit and Majumdar, Anirudha and Burchfiel, Benjamin and Dai, Hongkai and Simchowitz, Max},
  booktitle={CoRL 2024 Workshop on Mastering Robot Manipulation in a World of Abundant Data}
}

@misc{schulman2017proximalpolicyoptimizationalgorithms,
      title={Proximal Policy Optimization Algorithms}, 
      author={John Schulman and Filip Wolski and Prafulla Dhariwal and Alec Radford and Oleg Klimov},
      year={2017},
      eprint={1707.06347},
      archivePrefix={arXiv},
      primaryClass={cs.LG},
      url={https://arxiv.org/abs/1707.06347}, 
}

@article{shao2024deepseekmath,
  title={Deepseekmath: Pushing the limits of mathematical reasoning in open language models},
  author={Shao, Zhihong and Wang, Peiyi and Zhu, Qihao and Xu, Runxin and Song, Junxiao and Bi, Xiao and Zhang, Haowei and Zhang, Mingchuan and Li, YK and others},
  journal={arXiv preprint arXiv:2402.03300},
  year={2024}
}

@article{liu2025flow,
  title={Flow-grpo: Training flow matching models via online rl},
  author={Liu, Jie and Liu, Gongye and Liang, Jiajun and Li, Yangguang and Liu, Jiaheng and Wang, Xintao and Wan, Pengfei and Zhang, Di and Ouyang, Wanli},
  journal={arXiv preprint arXiv:2505.05470},
  year={2025}
}

@article{Qwen2.5-VL,
  title={Qwen2.5-VL Technical Report},
  author={Bai, Shuai and Chen, Keqin and Liu, Xuejing and Wang, Jialin and Ge, Wenbin and Song, Sibo and Dang, Kai and Wang, Peng and Wang, Shijie and Tang, Jun and Zhong, Humen and Zhu, Yuanzhi and Yang, Mingkun and Li, Zhaohai and Wan, Jianqiang and Wang, Pengfei and Ding, Wei and Fu, Zheren and Xu, Yiheng and Ye, Jiabo and Zhang, Xi and Xie, Tianbao and Cheng, Zesen and Zhang, Hang and Yang, Zhibo and Xu, Haiyang and Lin, Junyang},
  journal={arXiv preprint arXiv:2502.13923},
  year={2025}
}

@article{ye2025imgedit,
  title={Imgedit: A unified image editing dataset and benchmark},
  author={Ye, Yang and He, Xianyi and Li, Zongjian and Lin, Bin and Yuan, Shenghai and Yan, Zhiyuan and Hou, Bohan and Yuan, Li},
  journal={arXiv preprint arXiv:2505.20275},
  year={2025}
}
}

% % WARNING: do not forget to delete the supplementary pages from your submission 
\clearpage
\setcounter{page}{1}
\maketitlesupplementary
In this supplementary material, we present additional details on our method design in Section \ref{appendix:method}.
We also present additional experimental setups, metric design, and human validation of metric in Section \ref{appendix:experiments}.
Furthermore, we provide results of ablation experiments in Section \ref{appendix:ablation} to verify the effectiveness of our \textsc{MotionNFT} method.
Lastly, we visualize qualitative examples comparing our method to both open-source and closed-source commercial models in \ref{appendix:model-comparison}, highlighting the failure cases in these models and pointing towards future research direction.

\section{Additional Method Details}
\label{appendix:method}
\subsection{Preliminaries}
% One core methodological shift in the recent advancement of diffusion models is the change from Denoising Diffusion Probabilistic Models (DDPMs)~\citep{Rombach_2022_CVPR} to Flow Matching models (FMMs)~\citep{labs2025flux1kontextflowmatching}.
In T2I diffusion models, the forward noising process perturbs clean data $x_0$ from real distribution $\pi_0$ by adding a scheduled Gaussian noise $\epsilon \sim  \mathcal{N}(0, \text{I})$.
The model then learns to reverse the noise and output clean images.
The shift from Denoising Diffusion Probabilistic Models (DDPMs) to Flow Matching models (FMMs) is essentially a change in the prediction target of the model, from predicting the added noise itself (DDPM) to estimating a ``velocity field'' from the noise sample to the clean sample (FMM).

In mathematics formulation, FMMs define $z_t = \alpha_t x_0 + \sigma_t \epsilon$ to be a noisy interpolated latent at timestep $t$ between the initial clean $x_0$ and the noise $\epsilon$, where $\alpha_t$ and $\sigma_t$ defines the scheduled noise at $t$.
Then, for noisy sample $z_t$ and textual context $c$, a FMM $v_{\theta}$ is trained to directly approximate the target constant velocity field $v = \frac{\text{d}\alpha_t}{\text{d}t} x_0 + \frac{\text{d}\sigma_t}{\text{d}t} \epsilon$ by minimizing the objective:
$$
\mathcal{L}_{\text{FM}}(\theta) = \mathbb{E}_{t,x_0 \sim \pi_0, \epsilon \sim \mathcal{N}(0, \text{I})} \left[ \Vert v_{\theta}(\mathbf{x}_t, t, c) - v \Vert^2_2 \right].
$$
This velocity prediction allows for efficient inference by solving the deterministic Ordinary Differential Equation (ODE), $dz_t = v_{\theta}(z_t, t, c) dt$ for the forward process.

\subsection{Diffusion Negative-aware Finetuning (NFT)}
DiffusionNFT~\citep{zheng2025diffusionnft} aims at finding not only the ``positive velocity'' $v^{*}(x_t, t, c) = v^{+}(x_t, t, c)$ that the model learns to predict, but also identifying the ``negative velocity'' $v^{-}(x_t, t, c)$ component that the model should steer away from.
The training objective of DiffusionNFT is:
\begin{equation}
\label{eq:diffusion-nft}
\small
\begin{split}
    \mathcal{L}(\theta) = \mathbb{E}_{c,\,
          \pi^{old}(x_{0} \mid c),
          t} \;
    & r\big\|
        v_{\theta}^{+}(x_{t},c,t) - v
      \big\|_{2}^{2}
    \\
    & +
    (1-r)\,\big\|
        v_{\theta}^{-}(x_{t},c,t) - v
      \big\|_{2}^{2}
\end{split}
\end{equation}
Where $v$ is the target velocity, $v_{\theta}^{+}$ and $v_{\theta}^{-}$ are the implicit positive policy and implicit negative policy, defined as combinations of the old policy and current training policy: 
\begin{equation}
\small
\begin{split}
&v_{\theta}^{+}(x_{t},c,t)
:= (1-\beta)\,v^{old}(x_{t},c,t)
   + \beta\,v_{\theta}(x_{t},c,t), \\
&v_{\theta}^{-}(x_{t},c,t)
:= (1+\beta)\,v^{old}(x_{t},c,t)
   - \beta\,v_{\theta}(x_{t},c,t).
\end{split}
\end{equation}
Naturally, we need an optimal reward $r$ to accurately estimate the likelihood of the current action to fall into the ``positive'' subset of all samples.
However, real-world reward models might differ in score distributions and scales.
To this end, DiffusionNFT transforms raw rewards $r^{raw}$ into the optimality reward:
\begin{equation}
\scriptsize
r(x_{0},c) :=\frac{1}{2}
+
\frac{1}{2}
\operatorname{clip}
\!\left[
\frac{
    r^{\mathrm{raw}}(x_{0},c)
    - \mathbb{E}_{\pi^{old}(\cdot \mid c)}
      r^{\mathrm{raw}}(x_{0},c)
}{
    Z_{c}
}, -1, 1
\right]
\end{equation}
Where $Z_c$ is a normalizing factor (e.g. standard deviation of global rewards).

\subsection{MotionNFT: Motion-Aware Reward for NFT}
We propose a optical flow-based \textbf{motion-centric reward scoring framework} for our \textbf{MotionNFT} method to compute how closely model-predicted motion matches the ground-truth motion.
% We adopt an off-the-shelf model~\citep{xu2023unifying} for optical flow estimation.
Our reward scoring process is illustrated as follows:
% We employ an optical flow-based metric, leveraging a pre-trained UniMatch~\citep{xu2023unifying} model $\mathcal{F}$ to estimate the flow. 
% , we define the motion reward $\mathbf{r}_{m}(\mathbf{X})$ to quantify motion alignment with three loss terms that measures deviation

\subsubsection{Motion Calculation}
\myparagraph{Optical Flow Calculation}
Given two images $\mathbf{I}_0$ and $\mathbf{I}_1$, an optical flow estimation model~\citep{xu2023unifying} $\mathcal{F}$ quantifies motion flow between them with $\mathbf{V} = \mathcal{F}(\mathbf{I}_{\text{orig}}, \mathbf{I}_{\text{edited}}) \in \mathbb{R}^{H \times W \times 2}$,  where $\mathbf{V}$ is a vector field that represents the motion of each pixel with a 2D vector.
In our case, given an input triplet $\mathbf{X} = (\mathbf{I}_{\text{orig}}, \mathbf{I}_{\text{edited}}, \mathbf{I}_{\text{gt}})$ containing triplets of the original image $\mathbf{I}_{\text{orig}}$, the model-edited image $\mathbf{I}_{\text{edited}}$, and the ground truth image $\mathbf{I}_{\text{gt}}$,
% Given an input (pre-edit) image $\mathbf{I}_{\text{orig}}$, a ground-truth edited image $\mathbf{I}_{\text{gt}}$, and the model-edited image $\mathbf{I}_{\text{edited}}$, 
we first calculate the optical flow between the input image and the model-edited image: $\mathbf{V}_{\text{pred}} = \mathcal{F}(\mathbf{I}_{\text{orig}}, \mathbf{I}_{\text{pred}})$.
Then, we construct the motion reward $\mathbf{r}_{m}(\mathbf{X})$ to quantify the level of alignment between $\mathbf{V}_{\text{pred}}$ and the ground truth motion flow derived from the input and the ground-truth edited image $\mathbf{V}_{\text{gt}} = \mathcal{F}(\mathbf{I}_{\text{orig}}, \mathbf{I}_{\text{gt}})$ with three consistency terms: a \emph{motion magnitude consistency} term, a \emph{motion direction consistency} term, and an \emph{movement regularization} term.

\myparagraph{Flow normalization.}
Let $\mathbf{V}_{\text{pred}}(i,j) \in \mathbb{R}^2$ and
$\mathbf{V}_{\text{gt}}(i,j) \in \mathbb{R}^2$ denote the optical flow vectors at pixel $(i,j)$ for the model-edited and ground-truth edited images, respectively.
For an image of height $H$ and width $W$, we normalize the flows by the image diagonal to make the displacement magnitude comparable across resolutions:
\vspace{-0.4em}
\begin{equation*}
\scriptsize
    \tilde{\mathbf{V}}_{\text{pred}}(i,j) = \frac{\mathbf{V}_{\text{pred}}(i,j)}{\sqrt{H^2 + W^2}}, \tilde{\mathbf{V}}_{\text{gt}}(i,j)  = \frac{\mathbf{V}_{\text{gt}}(i,j)}{\sqrt{H^2 + W^2}}.
\end{equation*}
\vspace{-0.4em}

\subsubsection{Reward Calculation}
\myparagraph{Motion Magnitude Consistency Term.}
We first measure how closely the predicted flow magnitudes match the ground truth using a robust $\ell_1$ distance. Let $\mathbf{d}(i,j) = \tilde{\mathbf{V}}_{\text{pred}}(i,j) - \tilde{\mathbf{V}}_{\text{gt}}(i,j)$, magnitude deviation $\mathcal{D}_{\text{mag}}$ can be calculated as:
\vspace{-0.2em}
\begin{equation*}
\vspace{-0.5em}
\scriptsize
\mathcal{D}_{\text{mag}}
= \frac{1}{HW} \sum_{i=1}^{H} \sum_{j=1}^{W}
\bigl( \lVert \mathbf{d}(i,j) \rVert_{1} + \varepsilon \bigr)^{q},
\vspace{-0.4em}
\end{equation*}
where $\varepsilon > 0$ is a small constant used for numerical stability and the exponent \(q \in (0,1)\) enables a \emph{robust} variant of the $\ell_1$ distance
% Instead of the standard absolute difference, we raise each per-pixel residual to the power \(q\):
% \[
% (\lVert \mathbf{d}(i,j)\rVert_1 + \varepsilon)^q.
% \]
% Setting \(q < 1\) makes the penalty \emph{sub-linear}, which 
that suppresses the influence of large outliers in the flow field while still preserving sensitivity to semantically meaningful deviations.  
Empirically, we set \(q = 0.4\), which provides a stable trade-off between robustness and sensitivity for the motion-editing task.

\myparagraph{Motion Direction Consistency Term.}
We additionally measure directional alignment between the two flow fields, while focusing on regions with non-trivial motion.
Let:
\vspace{-0.3em}
\begin{equation*}
\scriptsize
m_{\text{gt}}(i,j) = \bigl\lVert \tilde{\mathbf{V}}_{\text{gt}}(i,j)\bigr\rVert_2, 
\;
m_{\text{pred}}(i,j) = \bigl\lVert \tilde{\mathbf{V}}_{\text{pred}}(i,j)\bigr\rVert_2,
\end{equation*}
\vspace{-0.3em}
and define unit flow directions:
\vspace{-0.3em}
\begin{equation*}
\scriptsize
\hat{\mathbf{v}}_{\text{gt}}(i,j)
= \frac{\tilde{\mathbf{V}}_{\text{gt}}(i,j)}{m_{\text{gt}}(i,j) + \varepsilon}, \;
\hat{\mathbf{v}}_{\text{pred}}(i,j)
= \frac{\tilde{\mathbf{V}}_{\text{pred}}(i,j)}{\lVert \tilde{\mathbf{V}}_{\text{pred}}(i,j)\rVert_2 + \varepsilon}.
\end{equation*}
\vspace{-0.3em}
We compute a cosine-based directional error per pixel:
\begin{equation*}
\vspace{-0.3em}
\scriptsize
\cos(i,j) = \hat{\mathbf{v}}_{\text{pred}}(i,j)^\top \hat{\mathbf{v}}_{\text{gt}}(i,j), 
\;
e_{\text{dir}}(i,j) = \tfrac{1}{2}\bigl(1 - \cos(i,j)\bigr),
\vspace{-0.3em}
\end{equation*}
and weight each pixel by the relative ground-truth motion magnitude:
\begin{equation*}
\vspace{-0.3em}
\scriptsize
w(i,j) = 
\frac{m_{\text{gt}}(i,j)}{\max_{u,v} m_{\text{gt}}(u,v) + \varepsilon}
\cdot \mathbf{1}\bigl[m_{\text{gt}}(i,j) > \tau_m\bigr],
\vspace{-0.3em}
\end{equation*}
where $\tau_m$ is a small motion threshold and $\mathbf{1}[\cdot]$ is the indicator function.
The directional misalignment $\mathcal{D}_{\text{dir}}$ can be calculated as:
\vspace{-0.4em}
\begin{equation*}
\scriptsize
\mathcal{D}_{\text{dir}} 
= \frac{\sum_{i,j} w(i,j)\, e_{\text{dir}}(i,j)}{\sum_{i,j} w(i,j) + \varepsilon}.
\end{equation*}
\vspace{-0.4em}

\myparagraph{Movement Regularization Term.}
While $\mathcal{D}_{\text{mag}}$ and $\mathcal{D}_{\text{dir}}$ encourage the predicted flow to match the ground-truth motion, they do not by themselves prevent the model from collapsing to a nearly static edit.
To discourage models from demonstrating this degeneration, we introduce a movement regularization term that compares the average motion magnitude of the predicted flow to that of the ground truth.
% Let
% \[
% m_{\text{gt}}(i,j) = \bigl\lVert \tilde{\mathbf{V}}_{\text{gt}}(i,j) \bigr\rVert_2, 
% \qquad
% m_{\text{pred}}(i,j) = \bigl\lVert \tilde{\mathbf{V}}_{\text{pred}}(i,j) \bigr\rVert_2,
% \]
We obtain the spatial means of ${m}_{\text{gt}}$ and ${m}_{\text{pred}}$:
\begin{equation*}
\scriptsize
\vspace{-0.4em}
\bar{m}_{\text{gt}} = \frac{1}{HW} \sum_{i,j} m_{\text{gt}}(i,j), 
\;
\bar{m}_{\text{pred}} = \frac{1}{HW} \sum_{i,j} m_{\text{pred}}(i,j),
\vspace{-0.4em}
\end{equation*}
and define the \emph{anti-identity} hinge term to be:
\begin{equation*}
\scriptsize
\vspace{-0.4em}
M_{\text{move}} 
= \max\bigl\{ 0,\ \tau + \tfrac{1}{2}\,\bar{m}_{\text{gt}} - \bar{m}_{\text{pred}} \bigr\},
\vspace{-0.3em}
\end{equation*}
where $\tau > 0$ is a small margin.
% This term is positive when the predicted motion is substantially smaller than the motion in the ground-truth edit (i.e., when the model is “under-moving”), and it vanishes once the predicted average motion is sufficiently close to or exceeds the ground-truth motion.
Intuitively, $M_{\text{move}}$ penalizes trivial edits that keep the image nearly identical to the input.

\myparagraph{Final: Motion-Centric Reward for training.}
Finally, we convert the optical flow-based alignment measure into a scalar reward for NFT training.
We combine the 3 terms to obtain:
\begin{equation*}
\scriptsize
\vspace{-0.4em}
\mathcal{D}_{\text{comb}} 
= \alpha\, \mathcal{D}_{\text{mag}} 
+ \beta\, \mathcal{D}_{\text{dir}} 
+ \lambda_{\text{move}}\, M_{\text{move}},
\vspace{-0.3em}
\end{equation*}
where $\alpha$, $\beta$, and $\lambda_{\text{move}}$ are hyper-parameters that balance the scales between magnitude and directional alignments, as well as assigning a small proportion to discouraging under-motion.
We normalize and clip the combined term:
\begin{equation*}
\scriptsize
\vspace{-0.4em}
\tilde{D}
= \mathrm{clip}\!\left(
\frac{\mathcal{D}_{\text{comb}} - \mathcal{D}_{\min}^*}{\mathcal{D}_{\max} - \mathcal{D}_{\min}^*},\, 0,\, 1
\right),
\vspace{-0.4em}
\end{equation*}
where $\mathcal{D}_{\min}^*$ is the lower bound of magnitude and directional terms calculated from a pair of duplicated inputs.
We then construct the continuous optical flow-based reward:
\vspace{-0.4em}
\begin{equation*}
\scriptsize
r_{\text{cont}} = 1 - \tilde{L} \in [0,1],
\vspace{-0.4em}
\end{equation*}
so that higher reward corresponds to better alignment with the ground-truth motion edit.
Finally, to approximate discrete human ratings of edited images following~\citep{luo2025editscore, lin2025uniworld}, we quantize the reward to 6 equally spaced levels:
\vspace{-0.4em}
\begin{equation*}
\scriptsize
\scriptsize
r_{\text{final}} = \frac{1}{5}\,\mathrm{round}\bigl(5\, r_{\text{cont}}\bigr) 
\in \{0.0,\, 0.2,\, 0.4,\, 0.6,\, 0.8,\, 1.0\},
\vspace{-0.4em}
\end{equation*}
which is the final scalar reward signal for MotionNFT.
During training, this raw reward score is further transformed to optimality rewards through group-wise normalization, and used to update the policy model $v_{\theta}$ by optimizing the DuffusionNFT objective in Equation \ref{eq:diffusion-nft}.

% Figure~\ref{fig:motionnft-pipeline} visualizes the reward scoring pipeline for MotionNFT.

% \paragraph{3. Anti-Identity (Movement) Hinge Term.}  
% We further incorporate a hinge term ($\mathcal{D}_{\text{move}}$) that aims at penalizing generated flow magnitudes that are too small relative to the ground truth, encouraging the model to generate the required movement $\Vert \mathbf{F}_{\text{gt}} \Vert$:
%     $$
%     \mathcal{L}_{\text{move}} = \max \left( 0, \tau + 0.5 \cdot \mathbb{E}[\Vert \mathbf{F}_{\text{gt}} \Vert] - \mathbb{E}[\Vert \mathbf{F}_{\text{pred}} \Vert] \right),
%     $$
%     with a margin $\tau$ (e.g., 0.01).

\section{Additional Evaluation Experiment Details}
\label{appendix:experiments}

\subsection{Hyperparameter Setting}
\subsubsection{Reward and Metric}
\myparagraph{MotionNFT Reward}
When calculating reward used for \textsc{MotionNFT}, we utilize three hyper-parameters to balance the three reward terms: $\mathcal{D}_{\text{comb}} 
= \alpha\, \mathcal{D}_{\text{mag}} 
+ \beta\, \mathcal{D}_{\text{dir}} 
+ \lambda_{\text{move}}\, M_{\text{move}}$
In our experiments, we set $\alpha = 0.7$, $\beta = 0.2$, and $\lambda_{\text{move}} = 0.1$.
Not only does this balance the scales between magnitude and directional alignments, as well as assigning a small proportion to discouraging under-motion.

\myparagraph{MAS Calculation}
When quantifying the MAS between model-edited images and ground truth targets, we punish degenerate cases where the predicted motion is nearly static compared to the ground-truth motion as a hard failure case and assign the minimum score $\text{MAS} = 0$.
This happens when $\frac{\mathbb{E}[m_{\text{pred}}]}{\mathbb{E}[m_{\text{gt}}]} < \rho_{\min}$), where $\rho_{\min}$ is a parameter determining how harsh the punishment threshold would be.
In our experiments, we set $\rho_{\min} = 0.01$.

\subsubsection{Model Training}
Following the training setup in ~\citet{lin2025uniworld}, we train all models with learning rate set to $3e-4$.
For FLUX.1 Kontext [Dev]~\citep{labs2025flux1kontextflowmatching} as base model, we report results for 300 steps.
For Qwen-Image-Edit~\citep{wu2025qwenimagetechnicalreport} as base model, we report results for 210 steps.
Due to computational limits, we set batch size to 2.
For NFT training, during sampling, we set sampling inference steps to 6, number of images per prompt to 8, and number of groups to 24; for training, we set KL loss' weight to $0.0001$ and guidance strength to 1.0.
For group filtering, we set the ban mean threshold to 0.9 and the standard deviation threshold to 0.05. 

\subsubsection{Model Inference}
During inference for Qwen-Image-Edit~\citep{wu2025qwenimagetechnicalreport} and the trained checkpoints, we set number of inference steps to 28, true cfg scale to 4.0, and guidance scale to 1.0.
For inference of FLUX.1 Kontext [Dev]~\citep{labs2025flux1kontextflowmatching} and its trained checkpoints, we set the same number of inference steps.
For inference of other open-sourced models, we follow the hyper-parameter setup in the official repositories. For UniWorld~\citep{lin2025uniworld}, we set number of inference steps to 25 and guidance scale to 3.5.
For AnyEdit~\citep{yu2024anyedit}, we set guidance scale to 3, number of inference steps to 100, and original image guidance scale to 3.
For UltraEdit~\citep{zhao2024ultraeditinstructionbasedfinegrainedimage}, we set number of inference steps to 50, image guidance scale to 1.5, and guidance scale to 7.5.
For Step-1X~\citep{liu2025step1x-edit}, we set number of inference steps to 28 and true cfg scale to 6.0.
For all other models, we set guidance scale to 3.5 and number of inference steps to 28.

\subsection{Human Evaluation for Generative Metrics}
\label{appendix:human-eval}
To evaluate the alignment between the MLLM-based generative metric and human judgment on motion editing quality, we conduct a human annotation study. 
Our annotators are a group of voluntary participants who are college-level or graduate-level students based in the United States. 
All annotators are proficient in English and have prior familiarity with AI research, ensuring that they understand the evaluation criteria and the purpose of the study. 
Prior to beginning, all annotators were informed that the anonymized results of their annotations may be used for research purposes only.

\subsubsection{Annotation Interface and Instructions}
We randomly sample 100 entries from \textsc{MotionEdit-Bench} for human evaluation, for which we further conduct random selection of outputs from 2 different models for comparison.
To ensure consistency, all annotators completed the same set of comparison tasks. 
Each annotation instance consists of five visual components: (1) the \textit{Input Image} to be edited, (2) the \textit{Ground Truth Edited Image} demonstrating the ideal motion change, (3) a \textit{Text Editing Instruction}, and (4–5) two model-generated edited outputs (labeled \textit{Model 1} and \textit{Model 2}). 
The annotators were asked to select which of the two model outputs better fulfill the requested motion edit, preserve the subject’s identity, and maintain overall visual coherence. 
Annotators were reminded that the Ground Truth serves as a reference only, not something to be matched pixel-wise.
They were encouraged to evaluate edits based on correctness of motion transformation and appearance preservation of the final image. 
If both outputs appear to be comparably good, annotators were instructed to select the one that is \textit{slightly better}. 
% An example annotation interface is shown in Fig.~\ref{fig:human-eval-interface}.

\myparagraph{Annotation Instruction.}
Before beginning annotation, participants read the following notice and instructions:

\begin{quote}
\vspace{-0.5em}
\textit{\textbf{Warning:} The set of model-synthesized images displayed below might contain explicit, sensitive, or biased content.}\\[4pt]
\textit{Thank you for being a human annotator for our study on the motion image editing task! By completing this form, you confirm voluntary participation in our research and agree to share your annotation data for research purposes only.}\\[4pt]
\textit{For each example, you will see: the Input Image, the Ground Truth Edited Image, an Editing Instruction, and two model-generated outputs. Your task is to determine which model output better follows the editing instruction \textbf{while preserving the identity and appearance of the subject}. Consider whether the edit is applied correctly, whether the subject remains consistent with the input, and whether the final image appears coherent and natural. You may consider the Ground Truth Image to be a ``reference answer'' of the ideal edit. If both outputs are similar in quality, choose the one you feel is slightly better.}
\vspace{-0.3em}
\end{quote}

\subsubsection{Human Evaluation Results.}
% \myparagraph{Human Evaluation Quality.}
Since all annotatros complete the same set of comparison tasks, each pair of model comparison was labeled by three independent annotators.
Inter-annotator agreement between all human annotators, as measured by Fleiss' $\kappa$, is $0.607$, indicating \textbf{good agreement} among human raters. 
The aggregated agreement between human annotators and decisions made by the overall generative metric (averaged over Fidelity, Preservation, and Coherence) achieves a \textbf{Fleiss' $\kappa$ score of $0.574$, similarly demonstrating substantial alignment between human judgment and our metric}. 
These results support the use of the MLLM-based generative evaluation metric as a practical and human-consistent measure of motion editing quality.

\section{Additional Evaluation Results}
% \subsection{Method Comparison}
% \ew{Results on balancing MLLM and Optical Flow-based Rewards}
% \begin{itemize}
%     \item 1.0 Optical Flow
%     \item 0.3 MLLM + 0.7 Optical Flow
%     \item 0.5 MLLM + 0.5 Optical Flow (reported)
%     \item 0.7 MLLM + 0.3 Optical Flow
%     \item 1.0 MLLM (reported)
% \end{itemize}
% % \ew{Maybe show steps vs. optical flow score \& score vs. MLLM score?}

\subsection{Ablation Studies}
\label{appendix:ablation}
\myparagraph{Balancing MLLM and Optical Flow-Based Rewards}
We investigate the optimal balancing strategy between our proposed optical flow-based motion alignment reward ($r_{\text{motion}}$) and the MLLM-based semantic reward ($r_{\text{mllm}}$) introduced in Uniworld-v2. 
Specifically, we adopt multi-objective reward NFT training with different weights for each reward.
Table 1 summarizes the editing performance on our \textsc{MotionEdit-Bench} across varying balancing weights. 
We observe that relying solely on the motion reward (1.0 * \textit{Motion}) leads to a performance degradation, indicating that geometric motion cues alone are insufficient for maintaining semantic fidelity. 
Conversely, while the pure MLLM reward (1.0 * \textit{MLLM}) provides a strong baseline, it is consistently outperformed by the combined approach.
The results demonstrate that the two objectives are complementary. 
The balanced configuration ($\lambda=0.5$) yields the highest performance across all metrics for both FLUX.1 Kontext [Dev]~\citep{labs2025flux1kontextflowmatching} and Qwen-Image-Edit~\citep{wu2025qwenimagetechnicalreport} backbones (achieving 4.25 and 4.72 Overall scores, respectively). 
This suggests that the optical flow reward effectively regularizes the MLLM guidance, improving motion alignment without compromising semantic coherence.

\begin{table}[h]
% \vspace{-0.3em}
\scriptsize
\centering
% 1 (Model) | 4 (MP) | 7 (ImgEdit) | 1 (Overall)  ==> 13 columns total
\begin{tabular}{l|p{0.04\textwidth}p{0.04\textwidth}p{0.04\textwidth}p{0.05\textwidth}}
\toprule
\multirow{2}*{\textbf{Model}} &
\multicolumn{4}{c}{\textbf{MotionEdit-Bench}}  \\
\cmidrule(lr){2-5}
  & \textbf{Ovl. $\uparrow$} 
 & \textbf{Fid. $\uparrow$} 
 & \textbf{Pre. $\uparrow$} & \textbf{Coh. $\uparrow$} \\ 
\midrule
FLUX.1 Kontext  & 3.84  & 3.89 & 3.79 & 3.83   \\
\textit{\;\;\; + 1.0 * Motion} &  3.60 & 3.62 & 3.60 & 3.59  \\
\textit{\;\;\; + 0.3 * MLLM + 0.7 * Motion} & 4.22 & 4.29   & 4.15 & 4.23  \\
\textit{\;\;\; + 0.7 * MLLM + 0.3 * Motion} & 4.16 & 4.23   & 4.08 & 4.16  \\
\textit{\;\;\; + 1.0 * MLLM} & 4.20    & 4.28 & 4.11  & 4.21 \\
\rowcolor{blue!14} \textit{\;\;\; \textbf{+ 0.5 * MLLM + 0.5 * Motion}}   & \textbf{4.25} & \textbf{4.33}  & \textbf{4.16}  & \textbf{4.25}   \\
\midrule
Qwen-Image-Edit   & 4.65 & 4.70 & 4.59 & 4.66 \\
\textit{\;\;\; + 1.0 * Motion} & 4.60 & 4.65 & 4.55 & 4.61 \\
\textit{\;\;\; + 0.3 * MLLM + 0.7 * Motion} & 4.72 & \textbf{4.81} & 4.61  & 4.74    \\
\textit{\;\;\; + 0.7 * MLLM + 0.3 * Motion} & 4.71 & 4.78 & 4.62 & 4.73  \\
\textit{\;\;\; + 1.0 * MLLM}   & 4.70  & 4.80 & 4.57 & 4.73 \\
\rowcolor{blue!14}  \textit{\;\;\; \textbf{+ 0.5 * MLLM + 0.5 * Motion}}  & \textbf{4.72}  & 4.79 & \textbf{4.63} & \textbf{4.74}  \\
\bottomrule
\end{tabular}
\vspace{-0.1em}
\caption{\label{tab:ablation-mllm-motionnft}Ablation experiments on different weights for balancing the MLLM-based reward proposed by ~\citep{lin2025uniworld} and our optical flow-based motion alignment reward. Results show that combining both rewards on a 0.5:0.5 scale achieves best performance, outperforming MLLM-only reward training.
}
% \vspace{-0.8em}
\end{table}

\begin{figure}[t]
\vspace{-0.3em}
 \centering
 \includegraphics[width=0.92\linewidth]{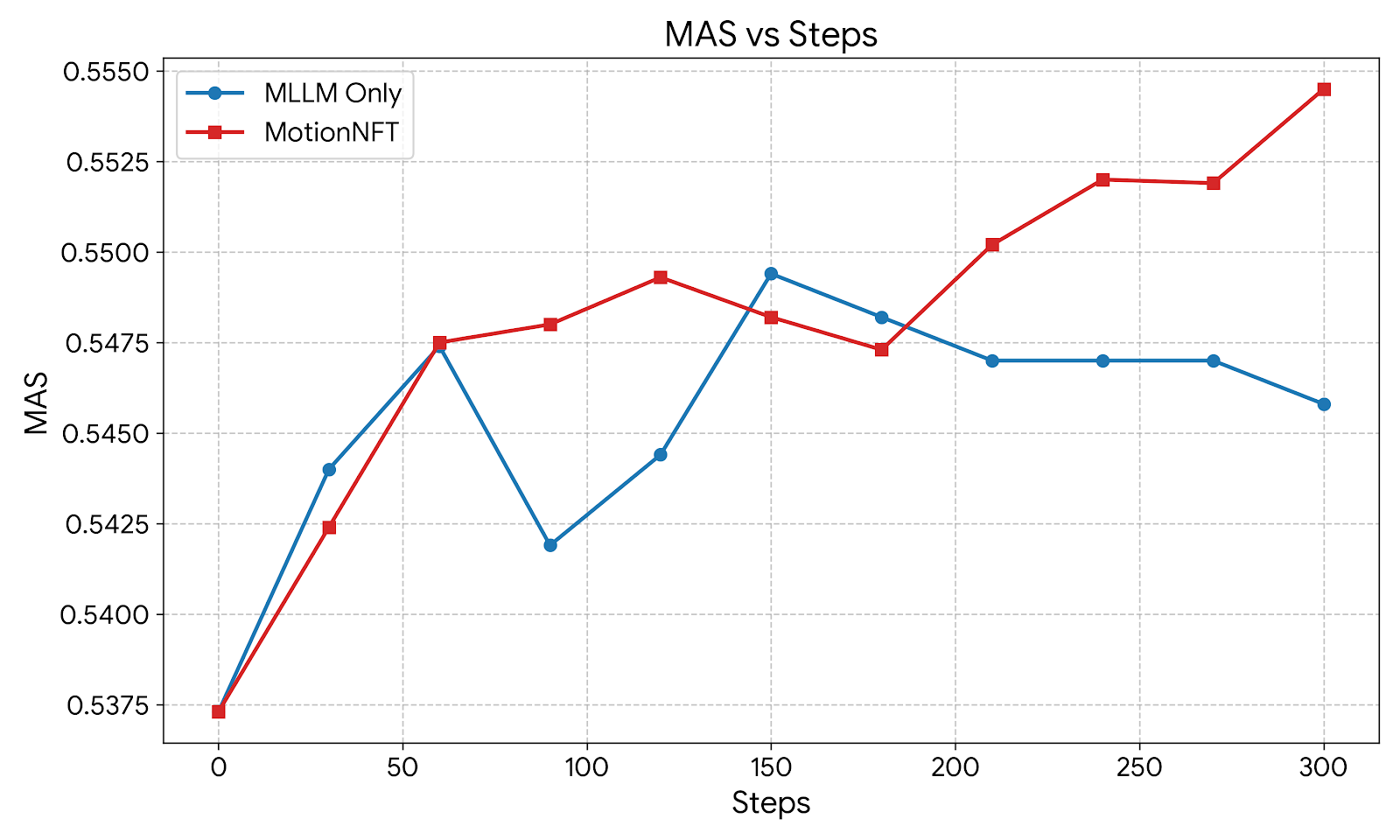}
 \vspace{-0.5em}
 \caption{\label{fig:mas-flux}MAS vs. Training Steps on FLUX.1 Kontext [Dev]~\citep{labs2025flux1kontextflowmatching}. MAS quantifies the fidelity of the generated motion by calculating the optical flow alignment (considering both magnitude and direction) between the model's edit and the ground truth target edit. While the MLLM-only baseline (blue) begins to regress after approximately 150 steps, MotionNFT (red) demonstrates steady improvement throughout training, ultimately achieving superior motion grounding by leveraging explicit motion guidance.}
 \vspace{0.2em}
\includegraphics[width=0.92\linewidth]{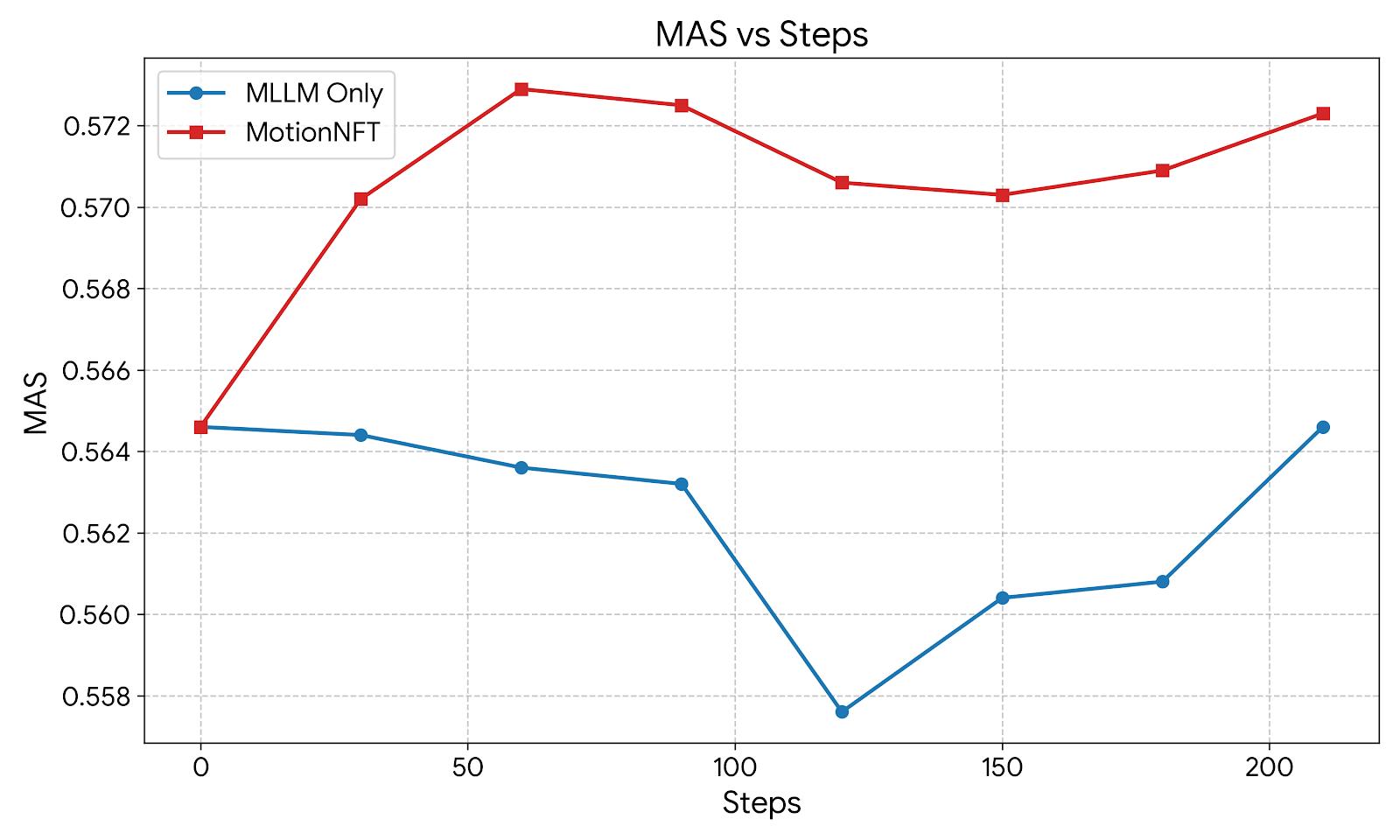}
 \vspace{-0.3em}
 \caption{\label{fig:mas-qwen}MAS vs. Training Steps on Qwen-Image-Edit~\citep{wu2025qwenimagetechnicalreport}. Results on other base models again shows that relying solely on semantic MLLM rewards leads to training regression in motion alignment. MotionNFT maintains prevents overfitting to semantic cues and achieving higher final MAS.}
\vspace{-1.0em}
\end{figure}

\myparagraph{MLLM-only Reward vs. MotionNFT}
Figures \ref{fig:mas-flux} and \ref{fig:mas-qwen} visualize the evolution of the Motion Alignment Score (MAS) during training for the MLLM-only reward in ~\citet{lin2025uniworld} and our \textsc{MotionNFT} reward.
As explained in previous sections, MAS utilizes optical flow to quantify magnitude and directional alignment level between model-edited motion and ground truth motion achieved in the target image.
We observe that relying solely on the MLLM-based semantic reward results in suboptimal motion alignment; for Qwen-Image-Edit (Fig. \ref{fig:mas-qwen}), the MAS even degrades significantly during the mid-training phase. 
In contrast, MotionNFT demonstrates robust and consistent improvement in MAS across both backbone models. 
By incorporating explicit motion guidance, our method prevents the model from overfitting to semantic cues at the expense of geometric accuracy, achieving a significantly higher final MAS.

\begin{figure*}[h]
\vspace{-0.6em}
 \centering
\includegraphics[width=0.9\linewidth]{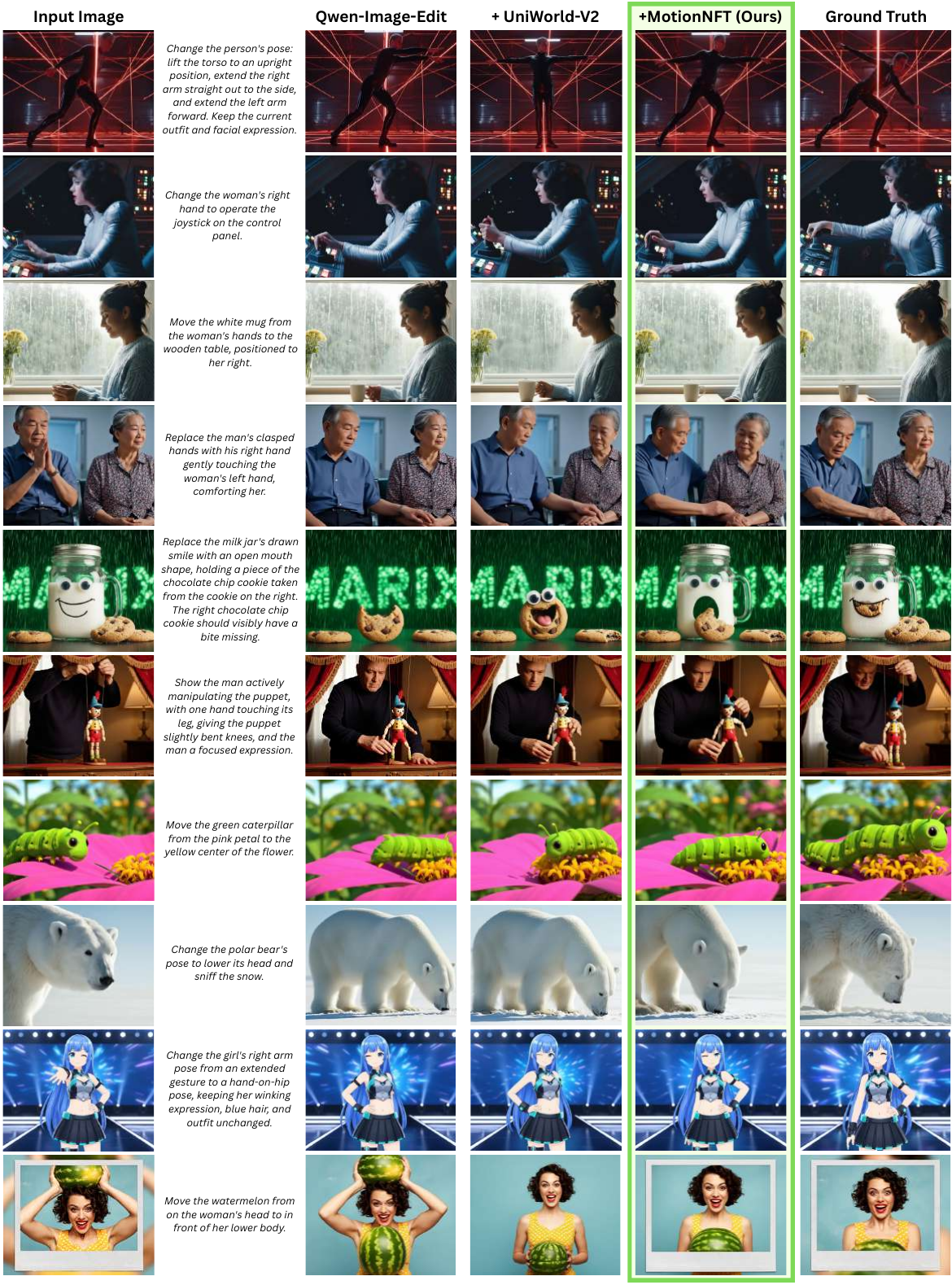}
 \vspace{-0.3em}
 \caption{\label{fig:suppl-example-qwen} Qualitative comparison of our method to Qwen-Image-Edit~\cite{wu2025qwenimagetechnicalreport} and the MLLM-only reward training in ~\citet{lin2025uniworld}. The base model frequently fails to demonstrate correct motion awareness for the edit (e.g. fail to move the subject's arms in the first row, and failing to displace the watermelon in the last row). While the MLLM-only approach improves semantic adherence, it often lacks geometric precision (e.g., caterpillar’s orientation in row 7). MotionNFT leverages optical flow to bridge this gap, achieving precise motion alignment and high fidelity to the editing instructions.}
\vspace{-0.8em}
\end{figure*}

\begin{figure*}[h]
\vspace{-0.6em}
 \centering
\includegraphics[width=0.9\linewidth]{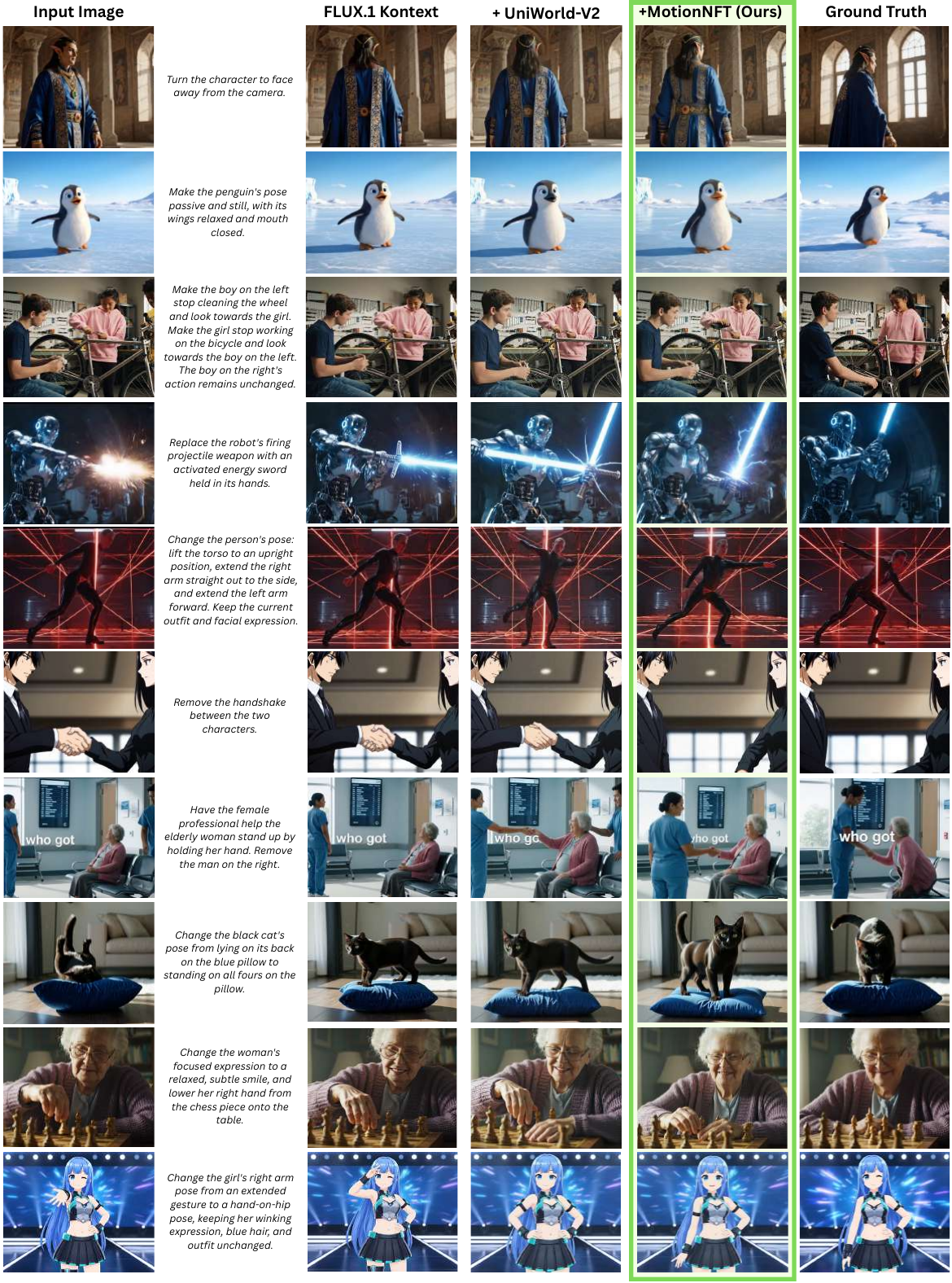}
 \vspace{-0.3em}
 \caption{\label{fig:suppl-example-flux}Qualitative comparison of our method to FLUX.1 Kontext [Dev]~\cite{labs2025flux1kontextflowmatching} and the MLLM-only reward training in ~\citet{lin2025uniworld}. The base model often exhibits editing inertia, failing to execute structural changes such as removing the handshake (row 6) or changing the subjects' directions (row 3). MLLM-only baseline also frequently hallucinates incorrect poses (e.g., the distorted limb placement in row 5) or fails to preserve subject identity (row 2). MotionNFT is able to interpret and execute complex motion edit instructions while preserving subject appearance and visual setting.}
\vspace{-0.8em}
\end{figure*}

\myparagraph{Qualitative Examples}
Figures \ref{fig:suppl-example-qwen} and \ref{fig:suppl-example-flux} compare MotionNFT against the base models (Qwen-Image-Edit~\citep{wu2025qwenimagetechnicalreport} and FLUX.1 Kontext [Dev]~\citep{labs2025flux1kontextflowmatching}) and their MLLM reward~\citep{lin2025uniworld}-guided counterparts. 
A recurring failure mode in the baselines is lack of ``motion awareness'', where the model fails to interpret and execute the desired motion subject, direction, and magnitude from the editing prompts.
For instance, in Figure \ref{fig:suppl-example-qwen} (Row 2), the Qwen baseline and the UniWorld-V2 baseline fails to correctly move the subject's right hand to operate the joystick, but rather placing both hands on it.
In Row 6, both baselines mistakenly flip the caterpillar's body direction when moving it towards the center of the flower.
In contrast, MotionNFT successfully executes both edits, matching the ground truth desired motion.

Additionally, we observe that another failure mode in baseline methods is the preservation of setting and subject identity.
In Figure \ref{fig:suppl-example-qwen} (Row 5), both baselines completely remove the milk jar despite it being a main subject in the image.
In the last row, both baselines remove the photo frame surrounding the woman that was in the original image, failing to preserve setting consistency.
Similarly, in Figure \ref{fig:suppl-example-flux} row 2, we observe that using \citep{lin2025uniworld}'s MLLM-only reward on FLUX.1 Kontext changes the penguin's beak in to a black color, failing to preserve its appearance while also not correctly performing the motion edit.
MotionNFT, on the other hand, achieves good preservation of subject's appearance and setting consistency.

\subsection{Model Comparison}
\label{appendix:model-comparison}
% \ew{Should we add Nano-Banana and GPT-Image-1?}
\subsubsection{Comparison with Open-Source Models}
\begin{figure*}[h]
\vspace{-0.6em}
 \centering
 \includegraphics[width=1.0\linewidth]{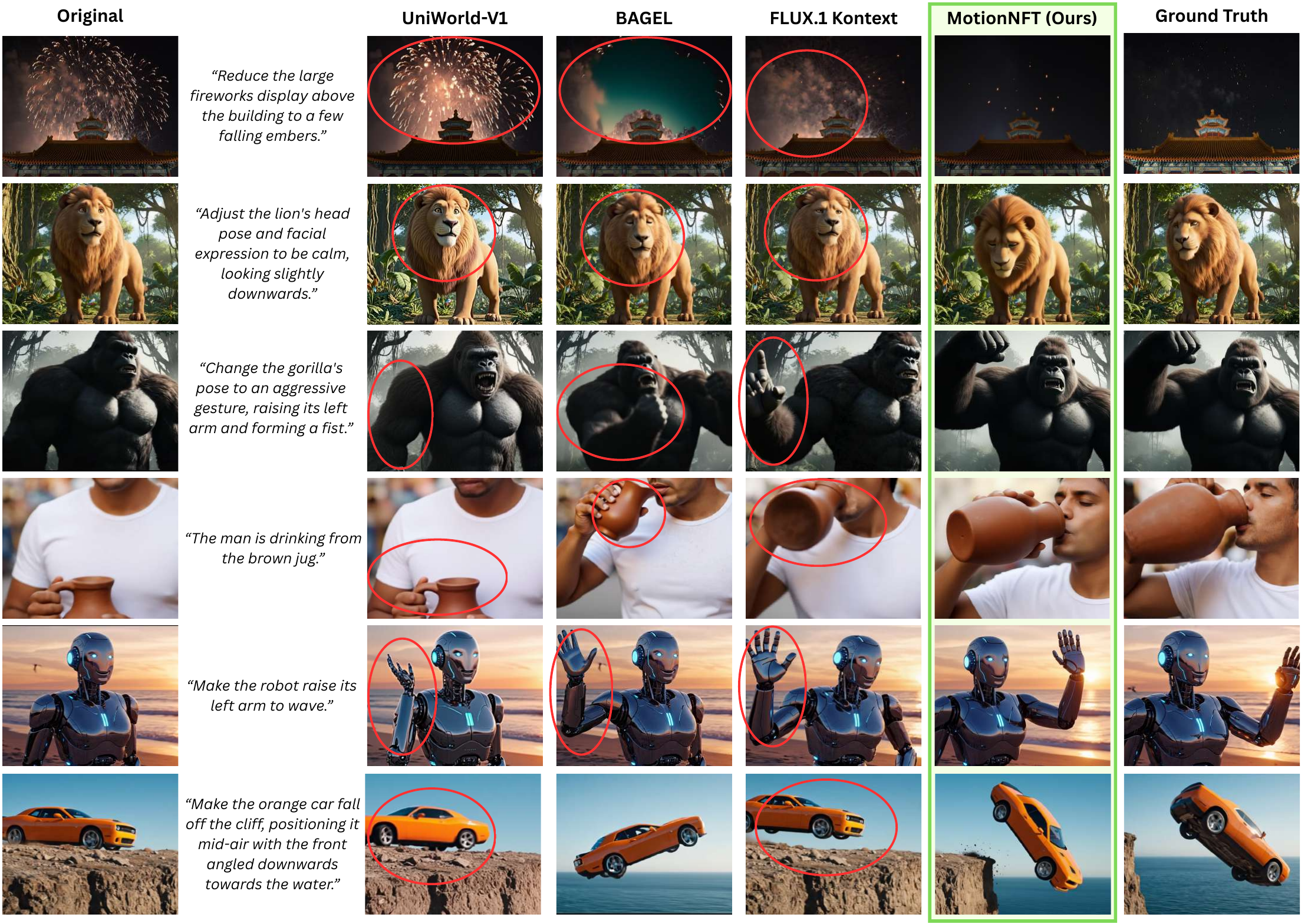}
 \vspace{-0.3em}
 \caption{\label{fig:fail-open}We compare MotionNFT against state-of-the-art baselines: UniWorld-V1~\citep{lin2025uniworld}, BAGEL~\citep{deng2025bagel}, and FLUX.1 Kontext [Dev]~\citep{labs2025flux1kontextflowmatching}. Red circles highlight failure regions. Baseline models exhibit different failure modes like editing inertia (e.g., failing to change the lion's pose in row 2), or motion misalignment (e.g., raising the robot's right arm instead of left arm in row 5). While baselines often struggle to execute challenging motion edits, MotionNFT achieves superior geometric grounding, accurately following semantic instructions and maintaining high motion fidelity to the ground truth.}
\vspace{-0.8em}
\end{figure*}

We compare MotionNFT against leading open-source editing models, including UniWorld-V1~\citep{lin2025uniworld}, BAGEL~\citep{deng2025bagel}, and FLUX.1 Kontext [Dev]~\citep{labs2025flux1kontextflowmatching}. Visual comparisons in Figure \ref{fig:fail-open} reveal that these baselines frequently struggle with precise motion controllability:
% Primary failure modes in the compared models include:
\begin{itemize}
    \item \textbf{Editing Inertia:} Existing models may fail to execute significant geometric transformations, defaulting to the original pose. For instance, in the "car cliff" scenario (Row 6), UniWorld-V1 fails to displace the vehicle, leaving it on the ledge with a flipped direction, while BAGEL and FLUX.1 lift the car but fail to capture the "downward angled" physics of the fall. Similarly, in the "lion" example (Row 2), all baselines fail to fully lower the head to the requested "looking downwards" pose, whereas MotionNFT achieves accurate alignment with the ground truth.
    \item \textbf{Motion Misalignment:} Existing models may fail to interpret and execute the subject part and direction of the motion change. For instance, in the gorilla example (Row 3), FLUX.1 Kontext fails to put the right hand into a fist. In the robot example (row 5), all baseline models fail to move the robot's left arm but move the right one instead. MotionNFT, on the other hand, performs the correct motion change on the correct subject part.
    \item \textbf{Structural Distortion:} When baselines do attempt large edits, they often introduce anatomical or semantic artifacts. In the "gorilla" example (Row 3), FLUX.1 Kontext distort the hand structure when attempting the ``fist'' gesture. In the jug drinking example (Row 4), the baselines leave residual artifacts that distorts the jug, while our method cleanly executes the edit without artifacts.
\end{itemize}

\subsubsection{Comparison with Closed-source Models}
\begin{figure*}[h]
\vspace{-0.6em}
 \centering
\includegraphics[width=1.0\linewidth]{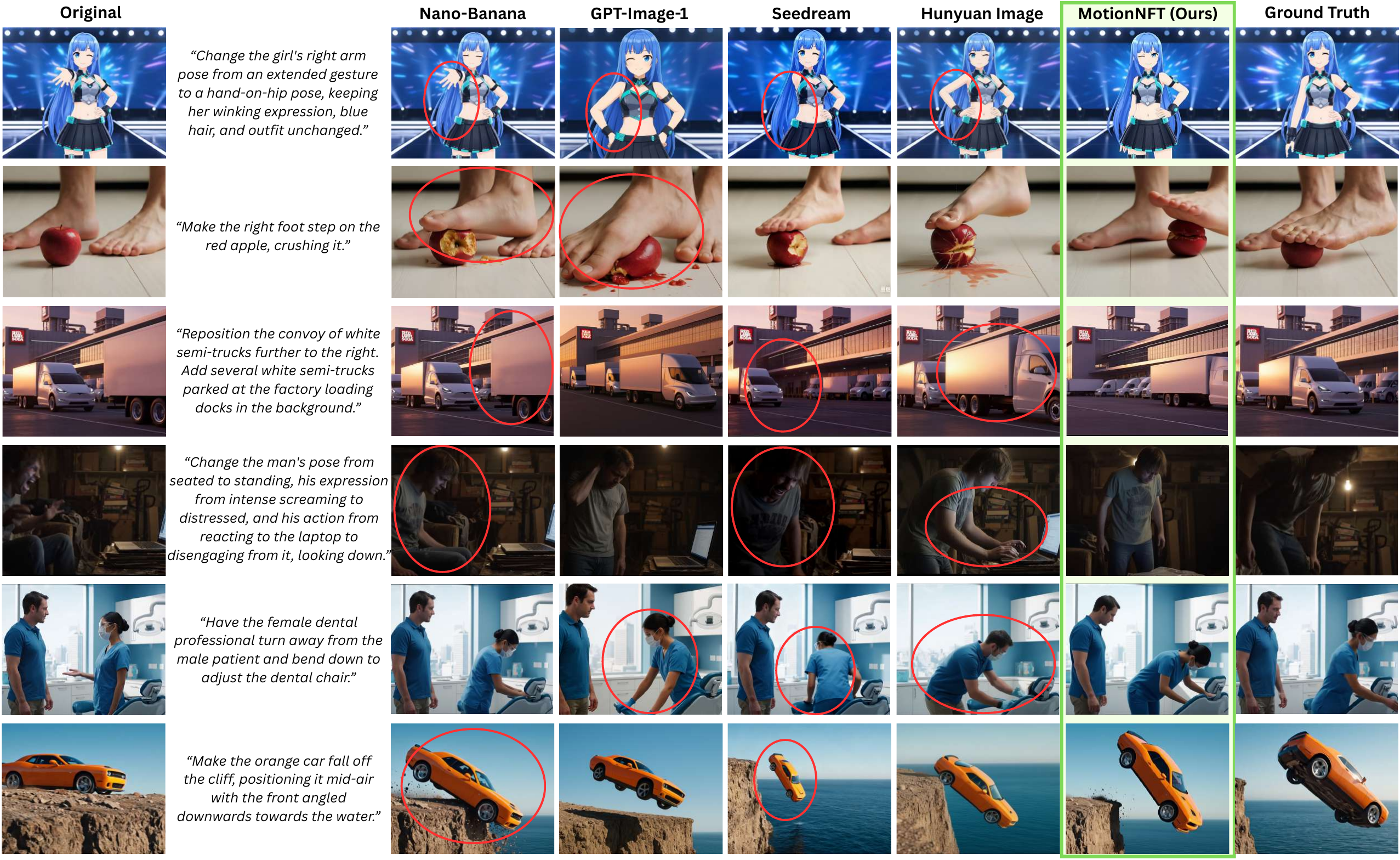}
 \vspace{-0.5em}
 \caption{\label{fig:fail-closed}We conduct selective case studies of MotionNFT against leading closed-source commercial baselines: Nano-Banana~\citep{nanobanana}, GPT-Image-1~\citep{openai2025imagegen}, Seedream~\citep{seedream2025seedream}, and Hunyuan Image~\citep{cao2025hunyuanimage}. Red circles highlight failure regions where baselines exhibit spatial inertia (e.g., failing to displace the car in the bottom row) or structural hallucination (e.g., generating an artifact ``foot'' in the second row). While commercial models generally maintain high visual quality, they frequently struggle to ground complex motion changes or maintain visual consistency. MotionNFT accurately follows these dynamic instructions, ensuring geometric alignment with the ground truth.}
  \vspace{0.3em}
 \includegraphics[width=1.0\linewidth]{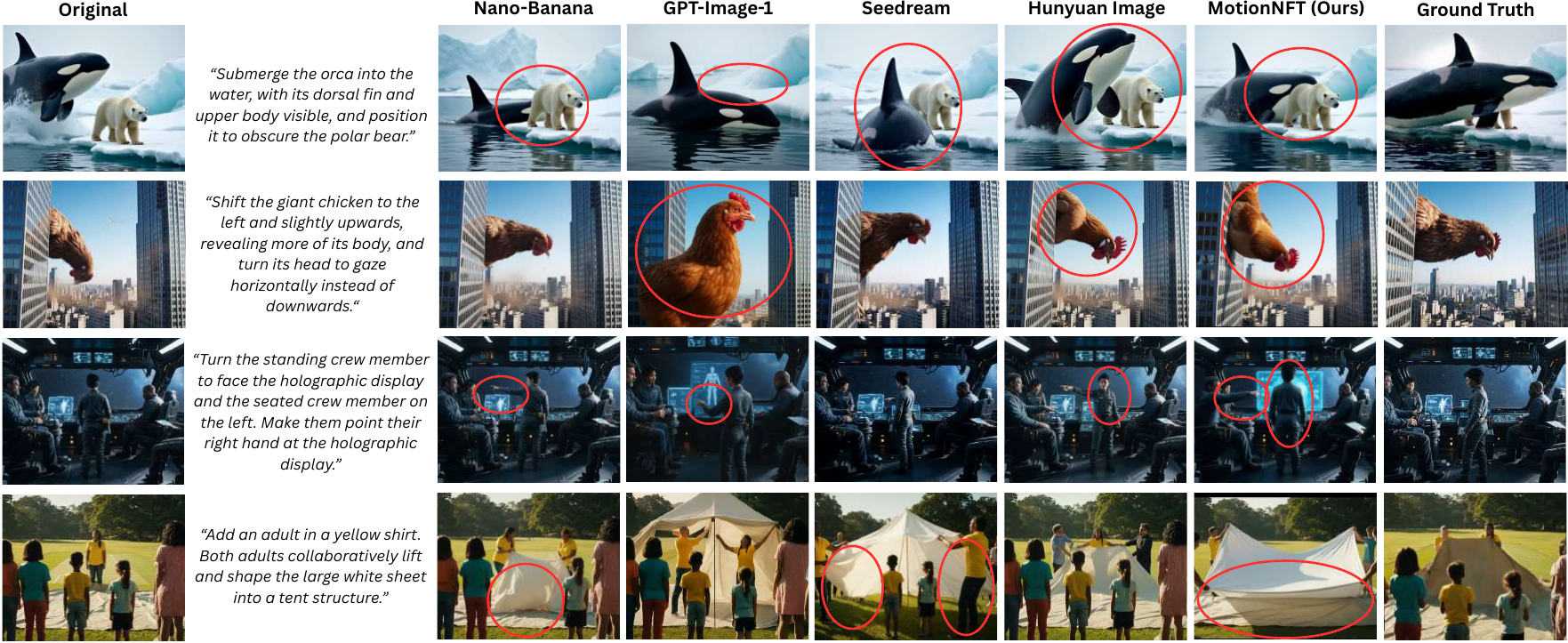}
 \vspace{-0.5em}
 \caption{\label{fig:fail-all}Additional failure cases of our model and closed-source commercial models. We observe that instructions involving multiple involving and non-involving subjects (e.g. the orca example in row 1, which requires complex 3D spatial edit) remain challenging for all evaluated methods. Current models, including ours and commercial baselines, struggle to correctly generate accurate and targeted motions on the correct subject part with the correct direction and magnitude in challenging scenarios.}
\vspace{-0.8em}
\end{figure*}

% \ew{Maybe say that we are on-par with several (2-3) closed-source models? (e.g. hunyuan image)}
We conduct selective case studies that compares MotionNFT with Qwen-Image-Edit as base model against leading closed-source commercial models, including Nano-Banana~\citep{nanobanana}, GPT-Image-1~\citep{openai2025imagegen}, Seedream~\citep{seedream2025seedream}, and Hunyuan Image~\citep{cao2025hunyuanimage}. 
As visualized in Figure \ref{fig:fail-closed}, these models still exhibit distinct failure modes for motion editing:
\begin{enumerate}
    \item \textbf{Semantic Hallucination and Structural Distortion:} When complex pose changes are required, baselines often introduce artifacts or unwanted semantic changes. In the "apple" example (Row 2), Nano-Banana introduces artifact by creating an additional ``feet'' that steps on the apple. MotionNFT avoids these structural collapses, successfully executing the editing instructions with high anatomical fidelity.
    \item \textbf{Motion Misalignment:} Even strong closed-source commercial models suffer from correctly identifying the subject part, direction, and magnitude of the motion change. For instance, in the anime girl example (row 1), Nano-banana demonstrates editing inertia where the motion of the subject remains the same, while GPT-Image 1, Seedream, and Hunyuan Image fail to execute the correct edit on the girl's arm.
\end{enumerate}

\subsubsection{Failure Analysis and Limitations}
While MotionNFT demonstrates robust performance across a wide range of editing cases, we observe specific scenarios where it, alongside leading commercial models, encounters difficulties. Figure \ref{fig:fail-all} illustrates these common failure modes that highlights persisting challenges:

\begin{itemize}
    \item \textbf{Multi-Subject Interactions:} Challenging editing settings with multiple involving and non-involving subjects in images pose a major challenge for existing models. For instance, in the orca example (row 1), all models fail to position the orca in front of the polar bear while executing the motion change to make it submerge in water. Similarly, in the crew member example (row 3), changing only the direction and the motion of one subject among a couple is difficult for existing models.
    \item \textbf{Identity Preservation} Existing models still struggle to preserve subjects and their identities in complex scenes. For instance, in the chicken example (row 2), 3 models fail to preserve the chicken's appearance. In the tent example (last row), models fail to preserve additional subjects in the scene not involved in the motion change. 
\end{itemize}

These cases suggest that future work incorporating stronger physics-based priors or motion guidance could further resolve the remaining challenges.

\subsection{Speed and Inference Cost}
MotionNFT is designed to be lightweight and computationally efficient. A key advantage of our method is that it can be seamlessly integrated with base models such as FLUX.1 Kontext Dev and Qwen-Image-Edit with no additional inference-time cost. All experiments were conducted on a single NVIDIA GPU. Using 28 sampling steps for a single entry, inference requires approximately 48 seconds with the FLUX.1 backbone and 98 seconds with Qwen-Image-Edit.
This confirms that MotionNFT enhances generation capabilities without compromising the speed or hardware requirements of the original models.

% \myparagraph{Win Rate}

% \begin{figure}[h!]
%  \centering
%  % \vspace{-4mm}
%  \includegraphics[width=0.9\linewidth]{figs/pairwise_winrates_heatmap.png}
%  \vspace{-1em}
%  \caption{Win Rate of different image editing models.}
%  \vspace{-0.3em}
%  \label{fig:motionnft-wins}
% % \vspace{-0.6em}
% \end{figure}

\end{document}